\documentclass[times,10pt,3p,review]{elsarticle}

\biboptions{sort&compress}

\usepackage{hhm}

\usepackage{comment}

\journal{Pattern Recognition} 

\begin{document}

\begin{frontmatter}

\title{Topology-Aware Active Learning on Graphs\tnoteref{t1}}
\tnotetext[t1]{Our source code is available at \url{https://github.com/hardiman-mostow/TopologyActiveLearning}.}

\author[1]{Harris Hardiman-Mostow\corref{cor1}}
\ead{hhm@math.ucla.edu}

\author[1]{Jack Mauro}
\ead{Jmauro@math.ucla.edu}

\author[1]{Adrien Weihs}
\ead{weihs@math.ucla.edu}

\author[1]{Andrea L. Bertozzi}
\ead{bertozzi@em.ucla.edu}

\cortext[cor1]{Corresponding author}
\affiliation[1]{organization={Department of Mathematics, University of California Los Angeles},
addressline={520 Portola Plaza},
postcode={90095},
city={Los Angeles},
country={USA}}

\begin{abstract}

Graph-based active learning seeks to improve classifier performance by iteratively selecting unlabeled points to query for labels. 
We propose a graph-topological approach to active learning that directly targets the core challenge of exploration versus exploitation under scarce label budgets. To guide exploration, we introduce a coreset construction algorithm based on Balanced Forman Curvature (BFC), which selects representative initial labels that reflect the graph’s cluster structure. 
Our method includes a data-driven stopping criterion that signals when the graph has been sufficiently explored. We further leverage BFC to dynamically trigger the shift from exploration to exploitation within active learning routines, replacing hand-tuned heuristics. To improve exploitation, we introduce a localized graph rewiring strategy that efficiently incorporates multiscale information around labeled nodes, enhancing label propagation while preserving sparsity. Experiments on benchmark image classification tasks show that our methods consistently outperform existing graph-based  baselines at low label rates in both sequential and batch active learning regimes.
\end{abstract}


\end{frontmatter}

\section{Introduction}

Supervised machine learning algorithms typically require vast amounts of labeled training data. However, in many real-world applications - such as medical imaging, remote sensing, or scientific data analysis - labels are expensive or impractical to obtain at scale, while unlabeled data is often abundant. To circumvent the need for large labeled training sets, \textit{semi-supervised learning} (SSL) aims to leverage both labeled and unlabeled data, harnessing the structure inherent in the dataset. 
One way to represent such structure is a \textit{similarity graph}, which encodes pairwise relationships between data points. 
A particularly successful family of SSL methods is \textit{graph-based semi-supervised learning} (GBSSL), which lifts the learning problem onto the similarity graph to exploit its geometry.
In the low-label regime, 
GBSSL exhibits strong performance in diverse applications such as image processing \cite{iscen2019label,sellars2022laplacenet}, remote sensing \cite{murphy2018unsupervised, brown2023utilizing, enwright2023deep}, data fusion \cite{iyer2020graph}, adversarial robustness \cite{wang2021graph}, few-shot learning \cite{liu2019learning, huang2021ptn}, and uncertainty quantification \cite{bertozziStuart}. 
Even in the era of deep learning, GBSSL remains an active area of research due to its compelling blend of empirical results, theoretical tractability, and interpretability (see e.g. \cite{weihs2025Hypergraphs,weihs2025HOHL} and references therein). 
In fact, 
GBSSL - when paired with \textit{active learning} (AL) - can beat state-of-the-art supervised neural network methods with a fraction of the labeled training data \cite{chapman_novel_2023}, and in some cases \textit{orders of magnitude} less training data \cite{chen2024cgap}.

When labels are scarce, the choice of which data points to label can significantly impact classifier performance. Active learning addresses this issue by iteratively building the labeled set
based on an \textit{acquisition function} that quantifies the ``usefulness'' of labeling each point in the unlabeled set. This enables careful selection of labeled points to maximize performance when labels are expensive to obtain. 
Queries can be made sequentially (i.e., one at a time, updating the underlying classifier after each query) or in batches, depending on the application \cite{settles2009active,chen2024cgap, chapman_novel_2023}.
In practice, labeling is typically performed by an expert oracle or ``human-in-the-loop''.

A key challenge in AL is the \textit{exploration–exploitation trade-off} \cite{miller2023poisson}: an active learner may use its queries to label unrepresented regions of the graph (explore), or label along the classifier's existing decision boundaries to refine and improve predictions (exploit).
Generally, exploration must precede exploitation; premature exploitation can lead to severe misclassifications by failing to sample entire regions of the input space. However, excessive exploration becomes label inefficient by failing to refine decision boundaries in later stages. Our approach addresses this trade-off in three novel ways:

\begin{itemize}
    \item improves exploration by selecting representative and diverse initial labels using the graph's topology, whereas previous work only considered path distances;
    \item strengthens exploitation by proposing a novel hypergraph-based method that localizes higher-order regularization near informative nodes;
    \item provides a graph-driven signal that guides when to transition from exploration to exploitation, replacing hand-tuned heuristics in prior work.
\end{itemize}

To achieve strong exploratory coverage, we  construct an effective \textit{coreset} - an initial set of labeled nodes selected purely from the graph structure and without any reliance on the classifier. 
A well-chosen coreset should provide sufficient coverage and reflect the diversity of the data distribution. 
This ensures that once exploitation begins, the acquisition function has enough information to make an effective query.
This is especially important when AL is done in batches \cite{chapman_novel_2023}; insufficient exploration can lead to catastrophic batch queries.

The main challenge of coreset selection is formalizing meaningful coverage of the graph. This problem was addressed in \cite{chapman_novel_2023}.
The authors introduced Dijkstras' Annulus Coreset (DAC),
which uses shortest-path distances to define metric balls on the graph, selecting nodes that collectively provide broad, quasi-uniform coverage. 
However, path distances do not account for the cluster structure of the graph (see Figure \ref{fig:blobs_illustration}). 
To address this, we introduce a graph curvature-based approach to coreset selection which leverages community structure in the graph and improves coverage beyond simple shortest-path distances. 

Once exploration yields a representative coreset, we shift focus to improving exploitation. 
To achieve this, we propose a novel approximation to multiscale Laplacian regularization, incrementally updating the graph Laplacian by incorporating higher-order structure only where it is most impactful: around labeled nodes. This localized rewiring strategy
preserves the sparsity of the graph while harnessing the benefits of multiscale regularization, achieving a favorable trade-off between accuracy and efficiency.  
Experiments demonstrate this approach significantly outperforms standard Laplace learning, while running over an order of magnitude faster than full multiscale methods.

\textbf{Our contributions:} This work introduces a novel active learning framework leveraging
\textit{Balanced Forman Curvature} (BFC) \cite{ricciBronstein} and multiscale graph Laplacian regularization \cite{Merkurjev,weihs2025HOHL,weihs2025Hypergraphs} to improve both exploration and exploitation. Specifically: 

\begin{enumerate}
    \item \textbf{Curvature coreset selection}: We develop a novel greedy algorithm, called \textit{Curvature Coreset} (CC, Algorithm \ref{alg:curvature}) that constructs a coreset via a minimax formulation of a BFC-based objective, promoting topologically-aware, diverse coverage across communities of the graph. Our method is hyperparameter-free and features a natural, graph-theoretic stopping criterion to determine when exploration is complete. 
    \item \textbf{Principled exploration-exploitation balance via BFC}: By integrating our BFC-based criterion into Poisson ReWeighted Laplace Learning with $\tau$-regularization (PWLL-$\tau$) \cite{miller2023poisson}, we introduce a data-driven mechanism that adaptively transitions from exploration to exploitation, replacing heuristic, fixed-schedule baselines used in prior work.
    \item \textbf{Localized Graph Rewiring for Label Propagation}: We enhance local graph structures around labeled nodes using a novel lightweight multiscale construction on a hypergraph, leading to significant performance gains at minimal computational cost.
    \item \textbf{Strong empirical validation}: We demonstrate consistent gains over strong baselines across image benchmark datasets in both coreset quality, computational efficiency, and downstream AL performance. 
\end{enumerate}

The remainder of the paper is structured as follows: in Section~\ref{sec:background}, we review the mathematical background relevant to our proposed methods; in Section~\ref{sec:methods}, we present our graph topology-based framework for active learning; in Section~\ref{sec:results}, we evaluate its effectiveness through numerical experiments; and in Section~\ref{sec:discussion} we conclude with a discussion of potential directions for future work.

\section{Background and Notation} \label{sec:background}

\subsection{Graph Laplacian Regularization}

This section presents the mathematical foundations for GBSSL generally, and the specific tools that underpin our methodology in Section~\ref{sec:methods}.

\subsubsection{Single-scale Graph Construction}\label{sec:background:graph_construct}

Given a dataset $\mathcal{X} := \{x_i\}_{i=1}^n \subset \mathbb{R}^d$ embedded in a feature space, GBSSL leverages a weighted \textit{similarity graph} that captures pairwise similarity between points. 
The similarity graph is represented using a weight matrix $W \in \mathbb{R}^{n \times n}$ with entries $w_{ij}$, where $w_{ij}$ is larger the more similar $x_i$ and $x_j$ are. The weights are typically computed using non-increasing kernel function $\eta: [0, \infty) \to [0, \infty)$, a scale parameter $\varepsilon > 0$, and setting the edge weights as
$w_{\varepsilon,ij} = \eta \left(\frac{\Vert x_i - x_j \Vert}{\varepsilon} \right)$.
This kind of similarity graph is known as a $\eps$-graph or a random geometric graph \cite{TutSpec}.
A canonical choice of $\eta$ is the indicator function $\eta(x) = \mathds{1}_{\{[0,1]\}}(x)$, so that $w_{\varepsilon,ij} \neq 0$ if and only if $\Vert x_i - x_j \Vert \leq \varepsilon$. 
The scale parameter $\varepsilon$ thus determines the radius of connectivity and plays a critical role in shaping the graph structure. 

An alternative graph construction is a $k$-Nearest Neighbor (kNN) graph \cite{TutSpec}. While $\varepsilon$-graphs and kNN graphs are closely related, $\varepsilon$-graphs are often more convenient for theoretical analysis, whereas kNN graphs are typically preferred in practice \cite{Slepcev, calder_poisson_2020}. We include the subscript $\varepsilon$ to specify $\varepsilon$-graph construction when necessary for theoretical precision, and omit it for simplicity when statements apply more broadly. Our experiments in Section \ref{sec:results} use kNN graphs.

To enable localized regularization in Section \ref{sec:methods:rewiring}, we define a local analogue of the weight matrix. Let \( \mathcal{S} \subset \{x_1, \dots, x_n\} \) be a subset of selected points.  
We define the localized weight matrix \( W_\varepsilon^{\mathcal{S}} \in \mathbb{R}^{n \times n} \) as
\[
(W_\varepsilon^{\mathcal{S}})_{ij} =
\begin{cases}
w_{\varepsilon, ij} & \text{if } x_i \in \mathcal{S} \text{ or } x_j \in \mathcal{S}, \\
0 & \text{otherwise},
\end{cases}
\]
i.e., we retain only edges adjacent to nodes in \( \mathcal{S} \) and set all other entries to zero. 

\subsubsection{Standard Graph Laplacian Regularization}\label{sec:background:GLR}

Many graph-based SSL methods are formulated as variational problems, where the goal is to find a labeling function $u$ defined as:
\begin{equation} \label{eq:intro:optimization}
u = \argmin_{v: \{x_i\}_{i=1}^n \to \mathbb{R}^p} J(v) + \Psi(v, \ell),
\end{equation}
where $J(v)$ is a regularization term promoting smoothness (according to the graph structure encoded in $W$), 
and $\Psi(v, \ell)$ enforces consistency with the known labels $\ell$. 
A canonical example is \textit{Laplace learning} \cite{LapRef} where the regularizer takes the form
\begin{equation} \label{eq:intro:laplace}
J(v) = \sum_{i,j=1}^n w_{ij} (v(x_i) - v(x_j))^2,
\end{equation}
promoting smoothness of the labeling function across connected nodes in the graph.
Given the  graph $G = (\mathcal{X}, W)$, the (unnormalized) \emph{graph Laplacian} is defined as
\[
L = D - W,
\]
where $D$ is the diagonal degree matrix with entries $D_{ii} = \sum_{j=1}^n w_{ij}$. 
Identifying a function $v:\{x_i\}_{i=1}^n \mapsto \mathbb{R}$ with a vector in $\mathbb{R}^n$, the regularizer in Laplace learning \eqref{eq:intro:laplace} can be rewritten as \[
J(v) = v^\top L v
\]
which directly shows that $L$ is symmetric and positive semi-definite \cite[Proposition 1]{TutSpec}. We note that normalized variants of the graph Laplacian, such as the symmetric or random-walk Laplacian, are also frequently used \cite{TutSpec}.
Laplace learning uses hard label constraints
\[
\Psi(v,\ell) = \begin{cases} 
0 & \text{if } v = \ell \text{ on labeled nodes}, \\
+\infty & \text{otherwise}.
\end{cases}
\]
In the case of binary labels, $\ell(x) \in \{0,1\}$), final labels $\bar{\ell}_i$ are assigned by thresholding, i.e. $\bar{\ell}(x_i) = \mathds{1}_{[0,0.5]}(u(x_i))$ for $i \in \mathcal{U}$. This formulation naturally extends to the multiclass setting with \( K \) classes. Let $S \in \mathbb{R}^{n \times K}$ be the one-hot label matrix (with rows $e_{\ell(x_i)}$ for $x_i \in \mathcal{L}$ where $e_j \in \mathbb{R}^K$ is the unit vector in $j$-th direction). Then, we use the regularizer \[
J(v) = \mathrm{Tr}(v^\top L v)
\]
for $v \in \mathbb{R}^{n \times K}$ 
and label-fidelity term
\[
\Psi(v,\ell) = \begin{cases} 
0 & \text{if } v_{i,:} = S_{i,:} \text{ on labeled nodes}, \\
+\infty & \text{otherwise}.
\end{cases}
\]
Here $M_{i,:}$ denotes the $i$-th row of the matrix $M$.
Once the optimal $u$ is computed, predicted class labels are given by:
\[
\bar{\ell}_i = \argmax_{k} u_{ik}.
\]
For notational simplicity, most of the remainder of the discussion will return to the binary classification case. When dealing with an $\eps$-graph, $G = (\{x_i\}_{i=1}^n, W_\varepsilon)$ 
we write $L_\eps$ for its Laplacian. Using $W_\varepsilon^{\mathcal{S}}$, we also define the corresponding localized graph Laplacian \( L_\varepsilon^{\mathcal{S}} \).

The Laplacian spectrum encodes rich information about the graph topology. For instance, the multiplicity of the zero eigenvalue of $L$ equals the number of connected components, and the corresponding eigenspace is spanned by the indicator vectors of those components \cite[Proposition 2]{TutSpec}. These spectral properties are central to many graph-based learning methods such as spectral clustering \cite{NgSpectral}, and have been studied extensively in the context of statistical consistence and convergence \cite{ GARCIATRILLOS2018239, hoffmann2020spectral}. In semi-supervised learning, graph Laplacian regularization with $J(v) = v^\top L_{\varepsilon} v$ penalizes the first-order variation of the labeling function over the graph \cite{Slepcev}. It serves as a discrete analogue of the squared gradient norm, encouraging smoothness by suppressing large differences between neighboring nodes.

We note that both the graph construction process and learning algorithms we consider here are distinct from the GNN setting \citep{rewiringMontufar}. GNNs typically operate on an \emph{a priori} given graph of relevant meta-data to the task (such as co-authorship in a citation graph), and train a neural network for various learning tasks. Conversely, we assume no graph structure of our data (and instead construct one over the dataset as described above), and the learning algorithms have no trainable parameters. However, we will later leverage a formulation of curvature proposed in the GNN community \cite{ricciBronstein} for use in our similarity graph-based SSL setting.

\subsection{Active Learning on Graphs} \label{sec:background:AL}

In graph-based AL, the goal is to identify valuable points from the unlabeled set $\mathcal{U} = \mathcal{X} \setminus \mathcal{L}$ to add to the labeled set $\mathcal{L} \subset \mathcal{X}$ to improve the current labeling function $u \in \mathbb{R}^{n \times K}$. The initial labeled set $\mathcal{L}_0$ is called a \textit{coreset} and subsequent querying is governed by an acquisition function 
\( \mathcal{A} : \mathcal{X} \to \mathbb{R}\). At each iteration, the algorithm queries the labels of the point

\[
x_{\text{acq}} = \argmax_{x \in \mathcal{U}} \mathcal{A}(x).
\]

See Algorithm~\ref{alg:activeLearning} for an overview of the AL loop. Among the most common choices for $\mathcal{A}$ is \textit{uncertainty sampling}. This method quantifies uncertainty in the classification by computing the \textit{margin}, or the gap between the predicted label $\bar{\ell}_i$ and the next most likely label:
\[
\mathrm{Margin}(x_i) = \bar{\ell}_i - \Bigl( \argmax_{k\neq \bar{\ell}_i} \, u_{ik} \Bigr)
\]

To fit the convention of acquiring the point with the maximum acquisition value, the acquisition function is then
\[\mathcal{A}_{unc}(x_i)  = 1 - \mathrm{Margin}(x_i).\]

The choice of $\mathcal{A}$ can also be motivated by a Bayesian formulation \cite{miller_model-change_2024}, where the regularizer $J$ defines a prior $\nu_1(v) \propto e^{-J(v)}$ and the fidelity term $\Psi$ defines a likelihood $\nu_2(\ell \mid v) \propto e^{-\Psi(v, \ell)}$. The posterior is then given by:
\[\mu(v \mid \ell) \propto e^{-J(v) - \Psi(v, \ell)},\]
and the MAP estimate coincides with the minimizer of the original variational problem \eqref{eq:intro:optimization}. 
This perspective naturally supports uncertainty quantification \cite{bertozziStuart}, enabling acquisition functions based on label variance \cite{ji12, ma2013sigma} or expected change in the underlying model \cite{miller_model-change_2024}. 

Many AL methods rely on the current classifier $u$ to guide label selection. However, when label information is scarce, this dependence can lead to degeneracies in the AL process, as the acquisition function cannot reliably distinguish which points are truly informative.
For example, uncertainty sampling - an ``exploitative'' active learner that selects points near the current decision boundary of $u$ - can perform poorly when the classifier is unaware of how many decision boundaries the dataset contains, over-refining some while ignoring others \cite{miller2023poisson}. Alternatively, one can design AL acquisition functions that are primarily ``explorative''. Examples include V-Opt \cite{ji12}, $\Sigma$-Opt \cite{ma2013sigma}, LAND \cite{murphy2018unsupervised}, and CAL \cite{cloninger2021cautious}. However, these methods can be much more expensive to compute \cite{miller2023poisson}; for example, V-Opt and $\Sigma$-Opt require computing and storing an $n \times n$ covariance matrix after each new label acquisition. 
They also lack the ability to exploit once exploration is complete, limiting the empirical performance and label efficiency. 

In contrast, coreset methods - such as DAC and the one proposed in this work - rely solely on the graph structure, which is sparse and only computed once, and have built-in stopping conditions that trigger once exploration is complete. Hence, to balance exploring and exploiting, one can use coreset selection to handle exploration before switching to unceratinty sampling-based exploitation. By first ensuring that the AL model has access to sufficiently informative labels via a coreset, we minimize the risk of such undesirable behaviors before the AL process begins, while maintaining computational and label efficiency.

An orthogonal approach to the ``coreset and exploitative AL'' paradigm is to design an acquisition function that can both explore and exploit. 
One such method is PWLL-$\tau$ \cite{miller2023poisson}, which we later describe in its own subsection. 
A key difference is that PWLL-$\tau$ assumes access to at least one label per class at initialization, while coreset-based algorithms make no such assumptions.
However, as a further extension of our coreset methodology, we will show that the transition from exploring to exploiting in PWLL-$\tau$ can be optimized with our same graph curvature-based approach.

Finally, thus far we have considered \textit{sequential} active learning, where each iteration only queries one point. \textit{Batch} active learning \cite{chapman_novel_2023}, conversely, selects a query set $\mathcal{Q}$ of size $B$, $|\mathcal{Q}|=:B>1$. This improves efficiency by requiring fewer computations of the acquisition function $\mathcal{A}$ (which  also requires recomputing the underlying classifier). It also lowers the human labeling time by allowing several experts to work in parallel on a batch of queries, which can be a bottleneck in practice. However, batch AL requires care in designing a batch policy that avoids querying several points with redundant information; naively selecting the top $B$ minimizers may result in several queries adjacent to one another. One such policy is LocalMax \cite{chapman_novel_2023}, which requires query points in a batch to have a larger acquisition value than any of their neighbors, ensuring some diversity in queries. Batch AL also requires better coreset exploration than sequential AL; poor initial coverage can compound errors since many points are queried at once. 
Our experiments in Section \ref{sec:results} demonstrate that our proposed coreset method consistently outperforms existing approaches in both sequential and batch AL settings.

\begin{algorithm}
\caption{Active Learning (Sequential)}\label{alg:activeLearning}
\begin{algorithmic}[1]
\Require Dataset $\mathcal{X}$, number of iterations $k$, acquisition function $\mathcal{A}(\cdot)$
\Ensure Optimized labeled set $\mathcal{L}_k$ and corresponding classifier $u$

\State Initialize coreset $\mathcal{L}_0 \subset \mathcal{X}$, set $\mathcal{U}_0 = \mathcal{X} \setminus \mathcal{L}_0$
\For{$i = 0$ to $k - 1$}
    \State Train classifier $u_i$ using labeled set $\mathcal{L}_i$
    \State Select query point: \[
        x_{\mathrm{acq}} = \argmax_{x \in \mathcal{U}_i} \mathcal{A}(x)
    \]
    \State Update labeled and unlabeled sets: \[
        \mathcal{L}_{i+1} = \mathcal{L}_i \cup \{x_{\mathrm{acq}}\}, \quad
        \mathcal{U}_{i+1} = \mathcal{U}_i \setminus \{x_{\mathrm{acq}}\}
    \]
\EndFor
\State Train final classifier $u$ using labeled set $\mathcal{L}_k$
\end{algorithmic}
\end{algorithm}

\subsubsection{Dijkstra's Annulus Coreset}

The \emph{Dijkstra's Annulus Coreset} (DAC) algorithm \cite{chapman_novel_2023} constructs a coreset by iteratively selecting nodes from a graph $G = (\mathcal{X}, W)$ under two constraints:
\begin{itemize}
    \item Separation: any two labeled nodes must be at least a graph distance $r$ apart;
    \item Coverage: every node in the graph must lie within distance $R$ of some labeled node.
\end{itemize}
Here, the graph distance $d_G$ is defined by the shortest-path metric (via Dijkstra’s algorithm).

Let $\mathcal{L}_0 \subset \{x_i\}_{i=1}^n$ denote the current set of labeled nodes and consider $r < R$. At each iteration, the algorithm computes:
\begin{itemize}
    \item the seen set, i.e. all nodes within distance $r$ of any labeled point:
    \begin{align*}
    \mathcal{S} &= \bigcup_{x \in \mathcal{L}_0} B_r(x), \text{where }  B_r(x) := \{ y \in \{x_i\}_{i=1}^n : d_G(x, y) < r \},
    \end{align*}
    \item and the candidate set, i.e. all nodes within distance R but outside the seen set:
    \[
    \mathcal{C} = \left( \bigcup_{x \in \mathcal{L}_0} B_R(x) \right) \setminus \mathcal{S}.
    \]
\end{itemize}

A new point $x^\ast$ is selected uniformly at random from $\mathcal{C}$ and added to $\mathcal{L}_0$. If $\mathcal{C} = \emptyset$ but $\mathcal{S} \neq \{x_i\}_{i=1}^n$, the algorithm samples $x^\ast$ uniformly from $\{x_i\}_{i=1}^n \setminus \mathcal{S}$ instead. After each selection, the sets are updated via:
\begin{align*}
\mathcal{L}_0 &\gets \mathcal{L}_0 \cup \{x^\ast\}, \;
\mathcal{S} \gets \mathcal{S} \cup B_r(x^\ast), \;
\mathcal{C} \gets \left( \mathcal{C} \cup B_R(x^\ast) \right) \setminus B_r(x^\ast).
\end{align*}
The process terminates once $\mathcal{S} = \{x_i\}_{i=1}^n$, ensuring complete coverage. The resulting set $\mathcal{L}_0$ serves as the DAC coreset.

In practice and for simplicity, $r$ is set to be $R/2$. The performance of AL with DAC is sensitive to the choice of radii $R$, which are fixed, heuristically selected, and do not adapt to the underlying graph structure. Even for a fixed $R$, the size of the coreset can vary significantly due to the stochasticity, which limits applicability in settings where there is a fixed label budget or high cost of labeling (see \ref{sec:app:dac_coreset_sizes}). Moreover, DAC proceeds until the entire graph is covered, which may be unnecessary and inefficient in practice. In Section~\ref{sec:methods:coreset}, we introduce a more adaptive and cluster-aware coreset selection method based on graph curvature, which improves both performance and label efficiency by leveraging the graph’s intrinsic topology.

\subsubsection{Poisson ReWeighted Laplace Learning with $\tau$-regularization}\label{sec:background:pwll}

An influential example of a jointly designed classifier-acquisition function pair is PWLL-$\tau$ \cite{miller2023poisson}. PWLL-$\tau$ introduces an additional regularization term into the variational problem \eqref{eq:intro:optimization}, designed to enforce decay of the labeling function $u$ away from labeled points:
\begin{align*}
u &= \argmin_{v : \{x_i\}_{i=1}^n \to \mathbb{R}^d} \sum_{i,j=1}^n \gamma(x_i) \gamma(x_j) w_{ij} \Vert v(x_i) - v(x_j) \Vert^2 + \tau \sum_{i\in \mathcal{U}} \Vert v(x_i) \Vert^2 \\
&\text{subject to} \quad v(x) = e_{\ell(x)} \quad \text{for } x \in \mathcal{L}. \notag
\end{align*}
Here, the reweighting function $\gamma$ is computed by solving the graph Poisson equation:
\begin{align*} 
&\sum_{j=1}^n w_{ij} \left( \gamma(x_i) - \gamma(x_j) \right) 
= \sum_{x_k \in \mathcal{L}} \delta_{ik} - \frac{1}{N}, 
\quad \text{for all } 1 \leq i \leq n,
\end{align*}
where $\delta_{ik} = 1$ if $i=k$, and $0$ otherwise. The corresponding acquisition function is given by $\mathcal{A}(x) = \Vert u(x)\Vert_2$, which, due to the $\tau$ term, tends to favor points far from the current labeled set - thus promoting exploration\footnote{The convention adopted in that work is to acquire the \textit{argmin}, not argmax, of $\mathcal{A}(x)$.}.

To gradually shift from exploration to exploitation, the authors introduce a geometric decay schedule for $\tau \to 0$ over multiple iterations. The schedule is defined by
\[
\tau_{n+1} = \mu \tau_n, \quad \text{with} \quad
\mu = \left( \frac{\epsilon}{\tau_0} \right)^{\frac{1}{2K}},
\]
where $\epsilon = 10^{-9}$, and $\tau_n$ is set to zero after $2K$ iterations. The parameter $K$ controls how quickly the method transitions from exploration to exploitation, and is set to the number of classes in the dataset. This decay strategy is somewhat rigid and ad hoc: while it ensures a transition from exploration to exploitation, it requires prior knowledge of the number of classes to set the parameter $K$. This makes the method dependent on an pre-selected hyperparameter, rather than label locations and graph structure, limiting its applicability to problems where class counts or distributions are unknown. Moreover, this value of $K$ is not optimal in general; when to switch from exploring to exploiting will depend on the topology of the graph. In Section~\ref{sec:method:explorationExploitation}, we introduce an alternative exploration-exploitation strategy that is adaptive and data-driven - leveraging the graph topology via curvature - and demonstrate that this strategy is more flexible and performant across different datasets with variable classification structures.

\subsection{Graph Curvature}

Ricci curvature is a fundamental object of study in differential geometry that measures whether local geodesics diverge (negative curvature), converge (positive), or stay parallel (zero). Analogues of curvature have been adapted to graphs \cite{ollivier2010survey} and studied extensively for oversmoothing and oversquashing \cite{ricciBronstein} in GNNs, due to their ability to identify problematic \textit{bottlenecks} in the graph. Graph curvature has also been used for pooling \cite{feng_graph_nodate} and community detection \cite{tian2025curvature}. Given an edge $x_i \sim x_j$ (that is, $w_{ij} > 0$), curvature is positive when $x_i$ and $x_j$ have mutual neighbors (cliques or triangles), zero 
when the edge is in a grid-like structure (4-cycles or squares), and negative otherwise (tree-like structure). 
The notion of curvature we consider is \textit{Balanced Forman Curvature} (BFC) \cite{ricciBronstein} for its conceptual and computational simplicity, and leave others for future lines of research. To define BFC, we first need a few helpful definitions, originally formulated in Topping et al. \cite{ricciBronstein}: 

\begin{definition}[Neighborhoods of $x_i \sim x_j$]\label{def:neighborhood}

We define the following terminology to describe the neighborhoods of nodes and edges:

\begin{enumerate}
    \item Let $N_1(i) = \{x_j \; | \; w_{ij} > 0 \}$ be the set of neighbors of $x_i$.
    \item $\Delta(i,j) := N_1(i) \cap N_1(j)$ is the set of mutual neighbors of $x_i$ and $x_j$, i.e., nodes forming triangles based at $x_i\sim x_j$. 
    \item $\Box^{i}(i,j) := \{x_k \in N_1(i)) \setminus N_1(j), \, x_k \neq x_j \; | \; \exists x_w \in (N_1(k) \cap N_1(j)) \setminus N_1(i) \}$. More simply, $\Box^{i}(i,j)$ is the set of neighbors of $x_i$ forming squares (4-cycles) traversing $x_i \sim x_j$ without diagonals inside.
    \item Finally, we let $\gamma_{\text{max}}$ be a correction factor counting the maximum number of 4-cycles based at $x_i \sim x_j$ which include a common node. 
\end{enumerate}
    
\end{definition}

We are now prepared to define BFC, which is the key ingredient in our graph-based coreset selection algorithm. 

\begin{definition}[Balanced Forman Curvature]\label{def:bfc}
For an edge $x_i \sim x_j$ in the edge set $E$ of a graph $G$, and letting $d_i$ be the degree of node $x_i$, we define the Balanced Forman Curvature as:
\begin{align*}
\text{Ric}(i,j) & := \\ & -2 + \frac{2}{d_i} + \frac{2}{d_j} + 2 \frac{|\Delta(i,j)|}{\text{max}\{d_i,d_j\}} + \frac{|\Delta(i,j)|}{\text{min}\{d_i,d_j\}} + \frac{(\gamma_{\text{max}})^{-1}}{\text{max}\{d_i,d_j\}}(|\Box^{i}(i,j)| + |\Box^{j}(i,j)|)
\end{align*}
    
\end{definition} 

Another interpretation of BFC - and a central idea motivating our method - is that $\text{Ric}(i,j)$ is negative when $x_i \sim x_j$ forms a bridge between two communities. Implicit in this observation is the data points $x_i$ and $x_j$ \textit{belong to two different communities}. This is the exact challenge of ensuring exploration in coreset selection: efficiently acquiring diverse points from all regions of the graph. 
In Section \ref{sec:methods:coreset}, we leverage this intuition behind BFC to design a coreset selection algorithm that ensures meaningful coverage of the graph.
We further show that BFC can \textit{measure} when exploration is complete, which can inform a stopping condition of our method (Section \ref{sec:method:stopcond}) and a more flexible, data-driven transition to exploitation in PWLL-$\tau$ (Section \ref{sec:method:explorationExploitation}).

\section{Methods} \label{sec:methods}

\subsection{Curvature Coreset Selection} \label{sec:methods:coreset}

\subsubsection{Algorithm Description}

As previously discussed, highly negative curvature on an edge $x_i \sim x_j$ indicates a ``bridge between clusters.''
In graph-based coreset selection, sampling from different clusters is key to building an effective and label-efficient training set. 
Hence, our algorithm iteratively builds the coreset by choosing nodes that exhibit high negative curvature with respect to the existing coreset nodes. This allows the algorithm to account for the cluster structure - not just path distances - when determining the coreset. 

\begin{algorithm}[!ht]
\caption{Curvature Coreset (CC)}\label{alg:curvature}
\begin{algorithmic}[1]
\Require Dataset $\mathcal{X}$, Adjacency matrix $A$, number of points to label $n$, optional reduction parameter $r$, optional stopping condition check \texttt{StopCond}.

\Ensure Labeled coreset $\mathcal{L}$ of length $n$.

\State Choose an $x \in \mathcal{X}$ uniformly at random.
\State Initialize $\mathcal{L} \gets \{x\}$
\State Initialize unlabeled set $\mathcal{U} \gets \mathcal{X} \setminus {x}$

\If{Using \texttt{StopCond}} \Comment{See Section \ref{sec:method:stopcond}}
\State Initialize streaming curvature values $C \gets \emptyset$
\EndIf

\If{$r$ is not \textbf{None}}
    \State Reduce candidate points $\mathcal{U} \gets \text{topDegree}(\mathcal{U},r)$ \Comment{See Remark \ref{rem:reduction}}
\EndIf

\For{$k$ = $1$ to $n$}

\State Compute $\text{Ric}(i,j)$ for each $x_i \in \mathcal{U}$ and $x_j \in \mathcal{L}$\Comment{Def. \ref{def:bfc}}

\State Compute $\hat{x} = \argmin_{x_i \in \mathcal{U}}\max_{x_j \in \mathcal{L}} \text{Ric}(i,j)$

\If{Using \texttt{StopCond}}
    \State $c = \min_{x_i \in \mathcal{U}}\max_{x_j \in \mathcal{L}} \text{Ric}(i,j)$
    \State $C \gets C \cup c$ \label{line:curv_hist}
    \If{\texttt{StopCond}(C) \textbf{is} \textbf{True}}
    \State $C \gets C \: \cup \: \{\hat{x}\}$
    \State $\mathcal{U} \gets \mathcal{U} \setminus {\hat{x}}$
    \State \textbf{Terminate Algorithm}
    \EndIf
\EndIf

\State $\mathcal{L} \gets \mathcal{L} \: \cup \: \{\hat{x}\}$

\State $\mathcal{U} \gets \mathcal{U} \setminus {\hat{x}}$

\EndFor

\end{algorithmic}
\end{algorithm}

Our Curvature-based Coreset selection algorithm (CC) is detailed in Algorithm \ref{alg:curvature}. At each iteration, we add an unlabeled point to the coreset via a minimax formulation of BFC computed between the current coreset and candidate unlabeled points. The algorithm can either be terminated once a user-specified number of points are labeled or our proposed stopping condition (Section \ref{sec:method:stopcond}) is triggered. 

\begin{remark}[Contrast to GNN Applications]
Since we are interested in using curvature to identify useful coreset \textit{nodes} (compared to the GNN literature, which are typically concerned with bottleneck \textit{edges}), we compute $\text{Ric}(i,j)$ regardless of whether the edge $x_i \sim x_j$ exists. This ensures the search space at each iteration can include the entire set of nodes, not just those adjacent to current coreset nodes (which, typically, would be a poor choice to add to the coreset). 
\end{remark}

\begin{remark}[Adjacency Matrix]
Our algorithm uses only the binary adjacency matrix $A$ of the graph (defined as $A_{ij} = 1$ if $W_{ij} > 0$, else $A_{ij} = 0$), not the weighted adjacency matrix $W$. While $W$ contains more information than $A$, we found suitable notions of weighted curvature lacking in the current literature, particularly from a computational efficiency perspective. Moreover, using $A$ was sufficient to achieve outstanding low-label rate classification results. Developing practical notions of curvature for weighted graphs is an interesting line of future research.
\end{remark}

\begin{remark}[Reduction Parameter]\label{rem:reduction}
    Inspecting Definition \ref{def:bfc} and Algorithm \ref{alg:curvature} reveals a bias toward high-degree nodes. The $\frac{2}{d_i}$ and $\frac{2}{d_j}$ terms encourage the algorithm to choose points in high density regions, which tend to be points that are at the center - in some sense ``representative'' - of a cluster. This synergizes nicely with exploitative active learning, where queries tend to fall along the more sparsely-populated decision boundary. Moreover, we can use this bias to speed up the algorithm by only considering the top $1/r$ fraction of nodes \textit{by degree} - reducing the search space on the graph by a factor of $r$ - without sacrificing accuracy (we demonstrate this and discuss further in \ref{sec:app:reduction}). This can be viewed as a method of \textit{pruning} or \textit{prefiltering}, which is common in the coreset literature \cite{zheng2023coveragecentric}.
\end{remark}

\begin{figure*}[ht]
    \centering
    \begin{subfigure}[t]{.25\linewidth}
    \centering
    \includegraphics[width=\linewidth]{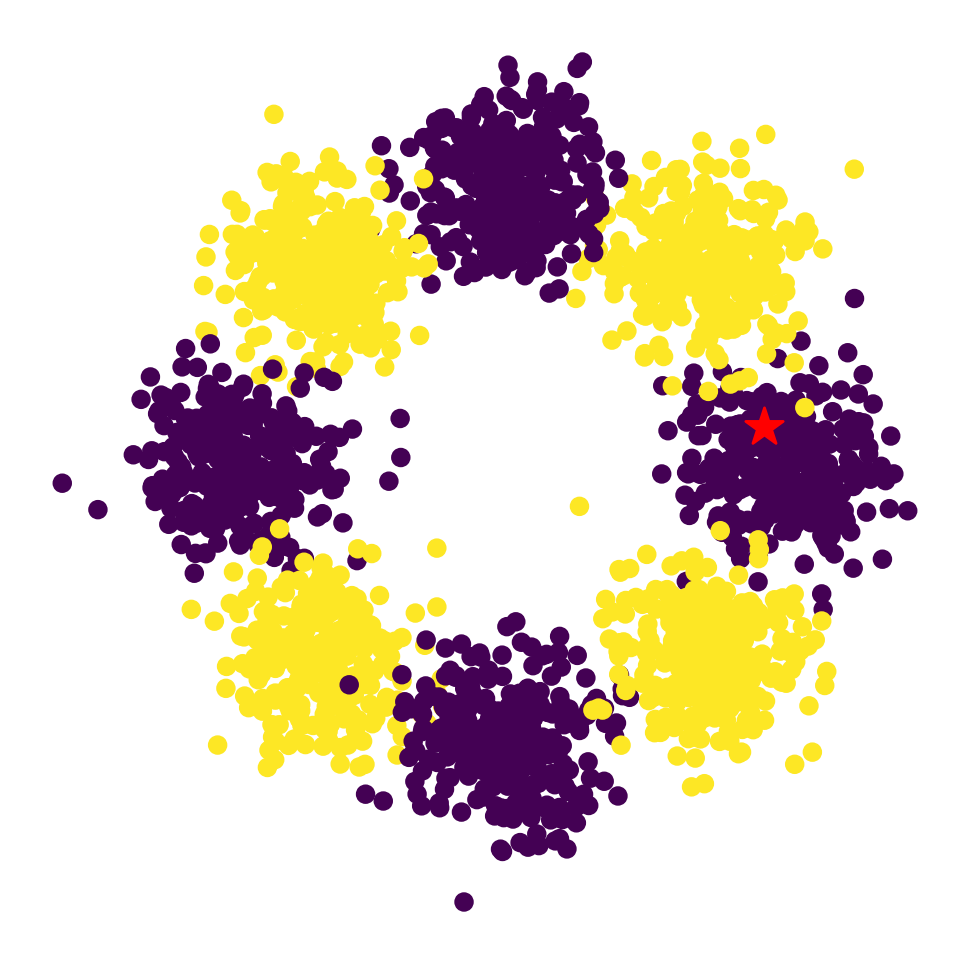}
    \caption{Curvature, Initial}
    \end{subfigure}
    \hfill
    \begin{subfigure}[t]{.25\linewidth}
    \centering
    \includegraphics[width=\linewidth]{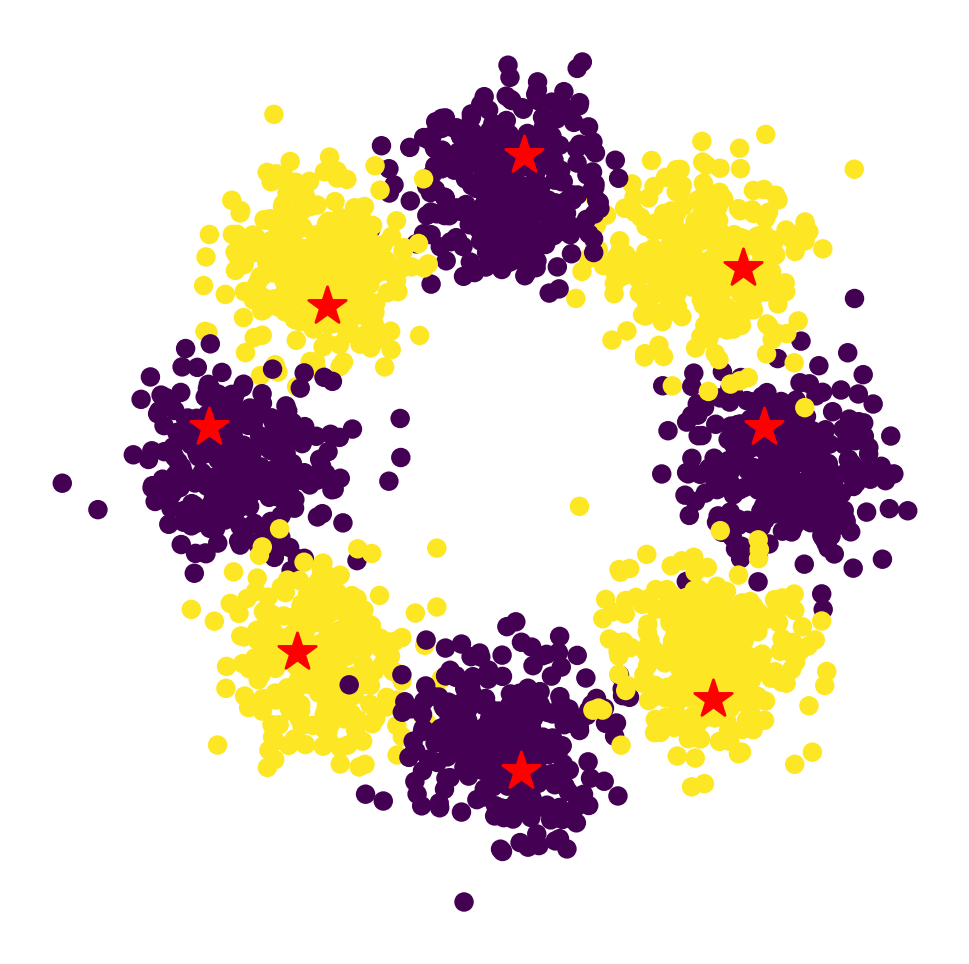}
    \caption{Curvature, Iter 8}
    \end{subfigure}
    \hfill
    \begin{subfigure}[t]{.25\linewidth}
    \centering
    \includegraphics[width=\linewidth]{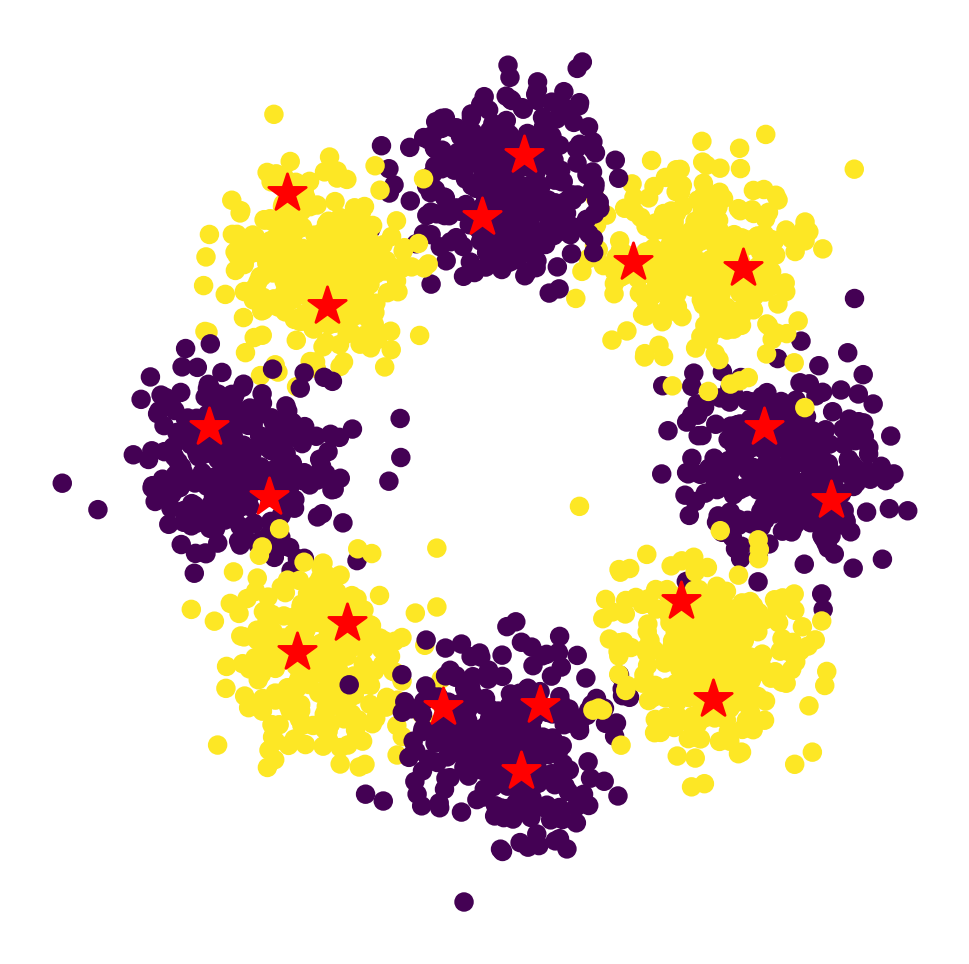}
    \caption{Curvature, Iter 17}
    \end{subfigure}
    \hfill
    \begin{subfigure}[t]{.25\linewidth}
    \centering
    \includegraphics[width=\linewidth]{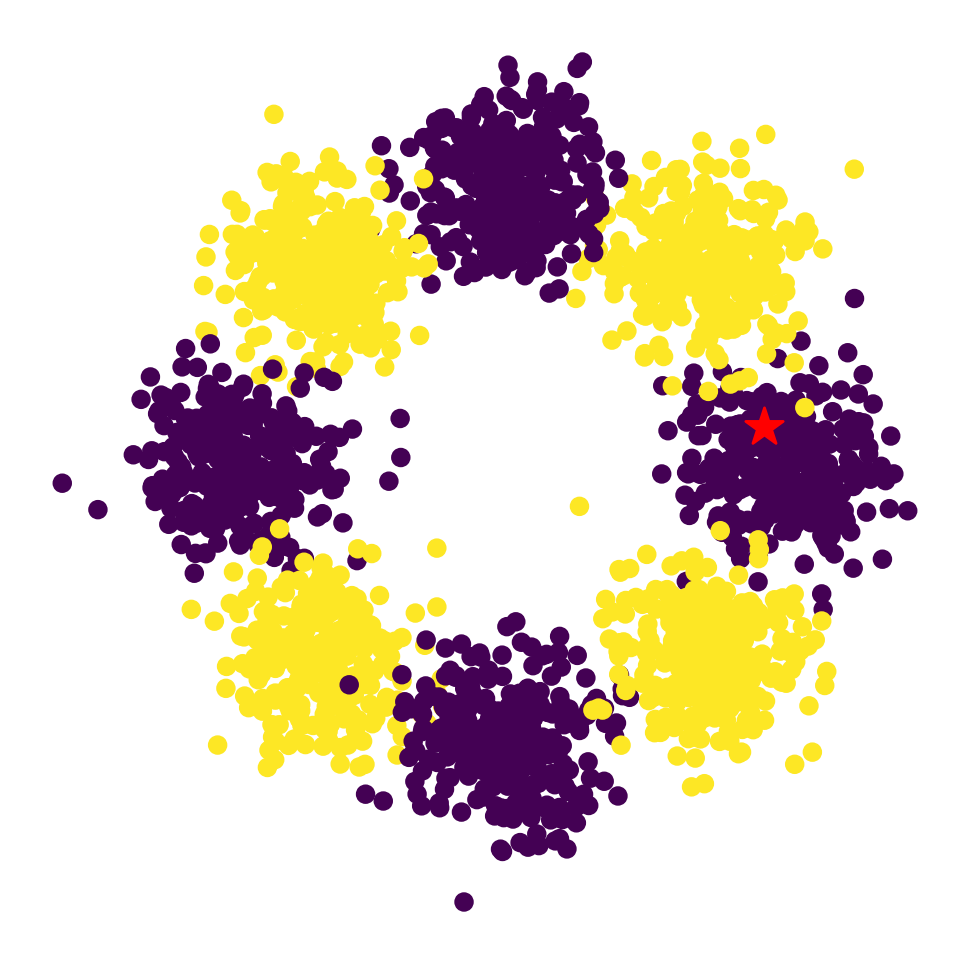}
    \caption{DAC, Initial}
    \end{subfigure}
    \hfill
    \begin{subfigure}[t]{.25\linewidth}
    \centering
    \includegraphics[width=\linewidth]{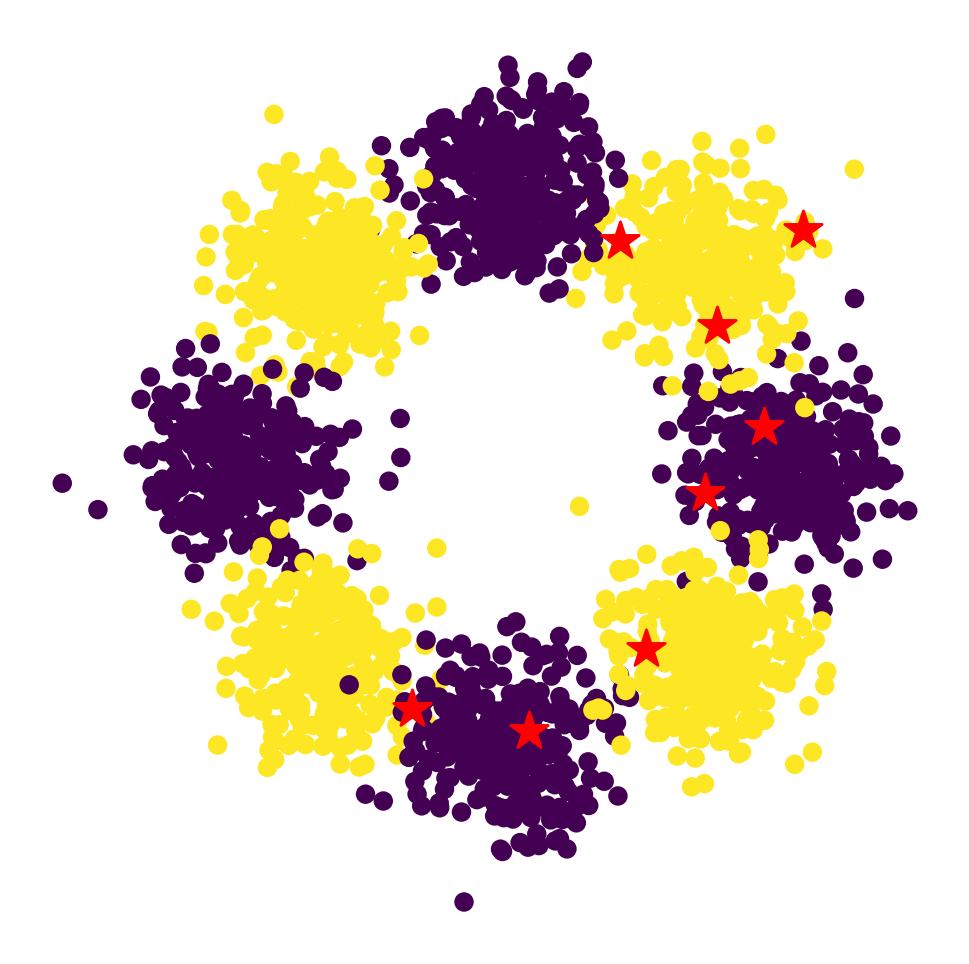}
    \caption{DAC, Iter 8}
    \end{subfigure}
    \hfill
    \begin{subfigure}[t]{.25\linewidth}
    \centering
    \includegraphics[width=\linewidth]{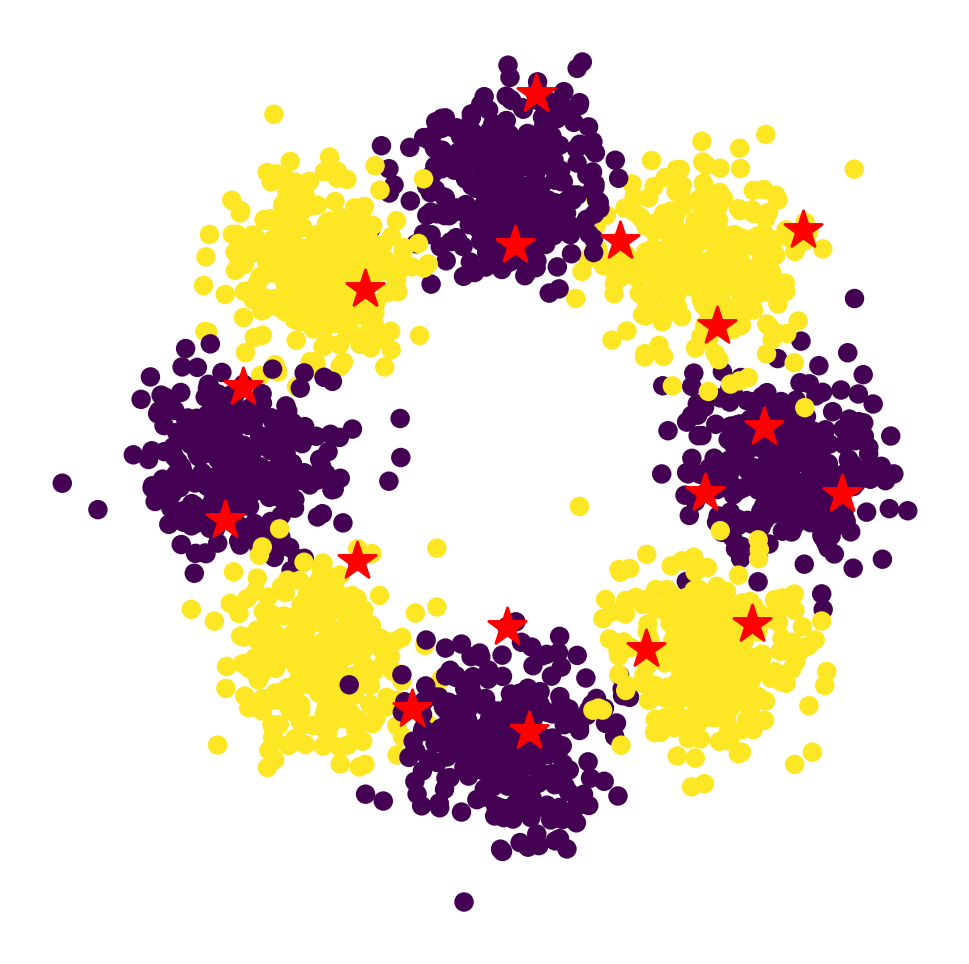}
    \caption{DAC, Iter 17}
    \end{subfigure}
    \caption{Coreset points chosen by CC and DAC at different iterations of coreset selection on the Blobs dataset. 
    }
    \label{fig:blobs_illustration}
\end{figure*}

We highlight and further motivate our method with an illustration on the ``Blobs'' dataset in Figure \ref{fig:blobs_illustration}. This dataset (previously used in \cite{miller2023poisson}) consists of eight Gaussian clusters with 300 points per cluster, centered at even spacings around the unit circle, with $\sigma = 0.17$, where the blobs are assigned alternating classes. We compare DAC with our proposed coreset algorithm, CC. DAC is slow to explore the dataset, requiring 17 iterations before each of the eight clusters has a point in the coreset. Meanwhile, CC samples from all eight clusters by iteration eight - over twice as fast as DAC. The curvature method is also biased toward higher-degree nodes, selecting points at the center of clusters, while DAC is random.

\subsubsection{Stopping Condition}\label{sec:method:stopcond}

\begin{figure}[ht]
    \centering
    \begin{subfigure}[t]{.4\linewidth}
    \centering
    \includegraphics[width=\linewidth]{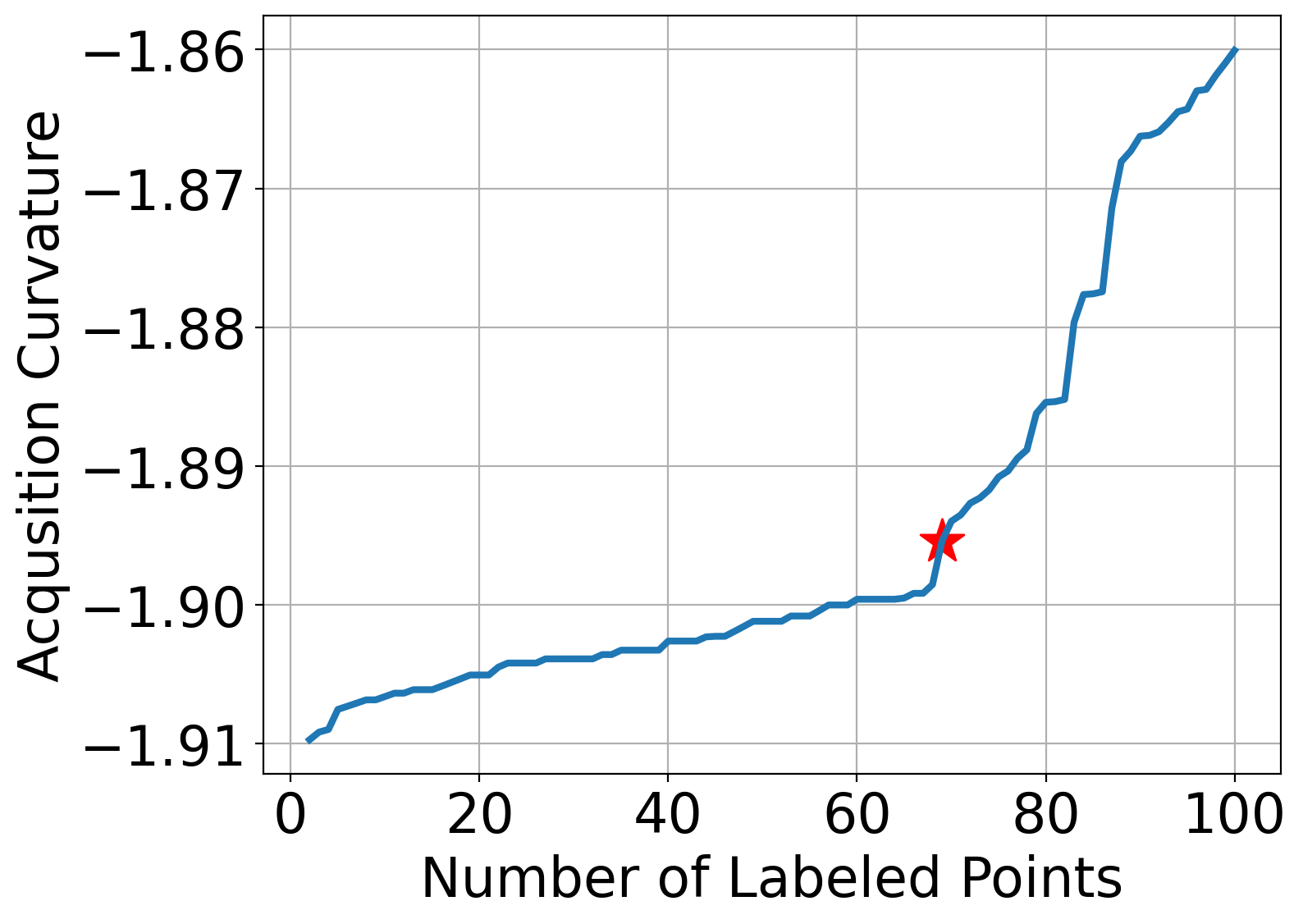}
    \caption{Value of $c$ (Line \ref{line:curv_hist}, Algorithm \ref{alg:curvature})}
    \end{subfigure}
    \hfill
    \begin{subfigure}[t]{.4\linewidth}
    \centering
    \includegraphics[width=\linewidth]{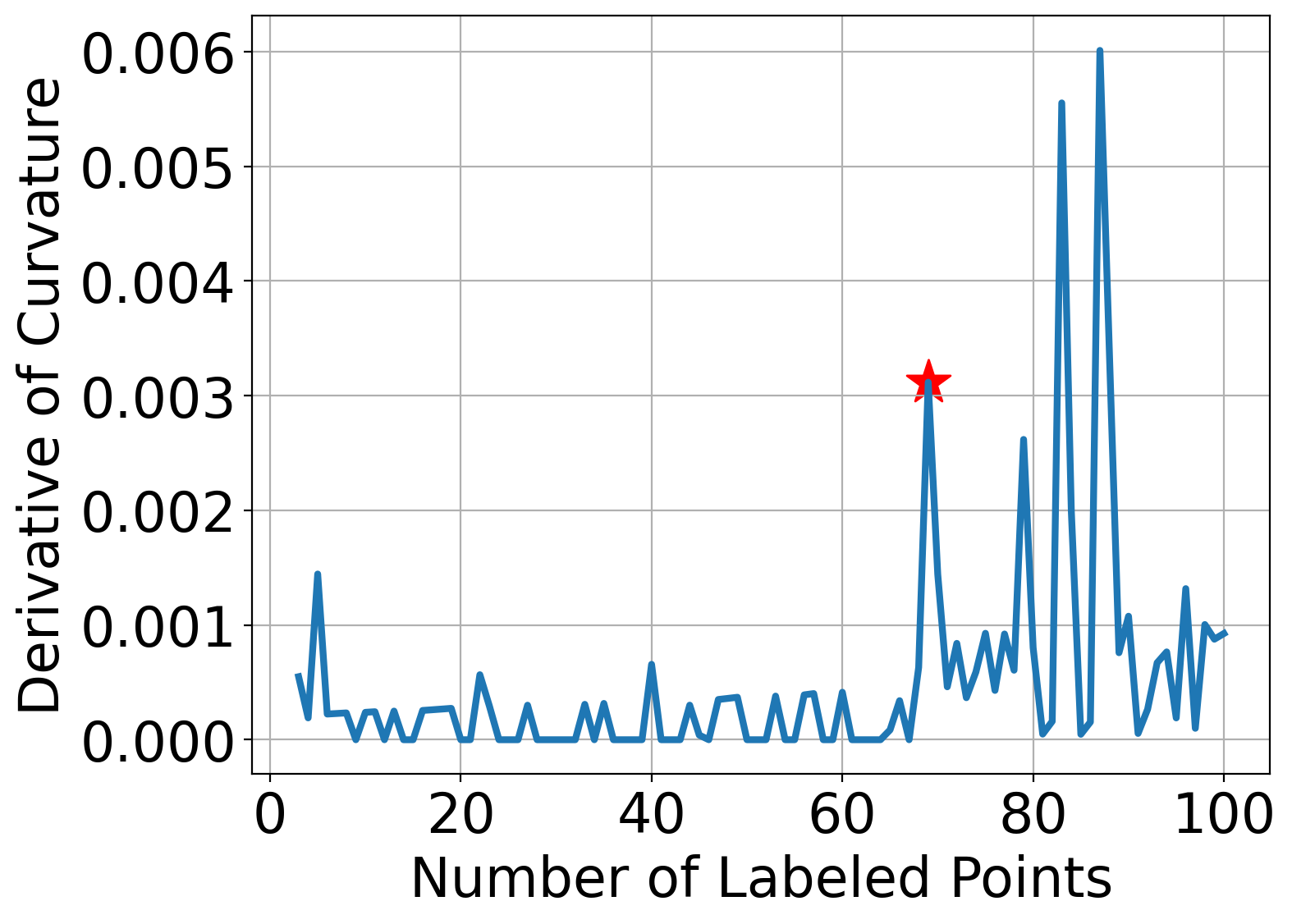}
    \caption{First-order difference of the sequence of $c$'s.}
    \end{subfigure}
    \caption{Illustration of our proposed stopping condition on MNIST. 
    The red star indicates when the online Z-score stopping condition triggers (Algorithm \ref{alg:curvature_stopping}), corresponding to the first large jump in curvature values among coreset points.}
    \label{fig:StopCond_illustration}
\end{figure}

Depending on the application, a stopping condition may be preferable to a fixed label budget for the coreset. For example, DAC stops when every node on the graph is within $R$ of a coreset point. This is often label-inefficient. Conversely, our stopping condition is online and data-driven, stopping when the underlying metric reflects that the graph is sufficiently explored.

Due to the minimax formulation of our coreset method, the value of curvature of the new coreset point between one iteration and the next will be nondecreasing. Figure \ref{fig:StopCond_illustration} illustrates that after sufficient exploration, there is a relatively large, sudden increase in the value of curvature between the acquired point and the existing coreset. This suggests the coreset has ``saturated'' the graph, and points toward a simple stopping condition: we stop once a large enough ``jump'' is made in BFC. To quantify this, we use a simple online anomaly detection algorithm based on rolling Z-scores. We record the discrete difference of successive values of BFC and compute a rolling mean and standard deviation of the last $N$ (we use $N$=20) differences. For each new sample (difference), we compute the Z-score, and the stopping condition triggers if the Z-score exceeds a threshold (we use $3$ standard deviations in our experiments). We provide pseudocode for this anomaly detector in Algorithm \ref{alg:curvature_stopping}.

\begin{algorithm}[t]
\caption{Online Z-Score Stopping Condition}\label{alg:curvature_stopping}
\begin{algorithmic}[1]
\Require Online sequence $C$ of curvature values (see Line \ref{line:curv_hist}, Algorithm \ref{alg:curvature}), Window size $N$, threshold $z_{\text{thresh}}$.
\Ensure Whether to end the coreset, \textbf{True} or \textbf{False}.
\State Initialize empty list $\mathcal{H} \gets \emptyset$
\For{each new data point $c_t \in C$}
    \State Append $c_t$ to $\mathcal{H}$
    \If{$|\mathcal{H}| > N$}
        \State Remove oldest element from $\mathcal{H}$ 
    \EndIf
    \If{$|\mathcal{H}| < N$}
        \State \textbf{continue} \Comment{Not enough data yet}
    \EndIf
    \State Compute mean $\mu \gets \text{mean}(\mathcal{H})$
    \State Compute std $\sigma \gets \text{std}(\mathcal{H})$
    \State Compute $z \gets |c_t - \mu| / \sigma$
    \If{$z > z_{\text{thresh}}$}
        \State \textbf{Return} \textbf{True} and end the coreset
    \EndIf
\EndFor
\State \textbf{Return False}
\end{algorithmic}
\end{algorithm}

\subsection{Principled Exploration–Exploitation Balance via BFC} \label{sec:method:explorationExploitation}

As previously discussed, PWLL-$\tau$ has a set decay schedule for $\tau$ which controls the tradeoff between exploration and exploitation. We instead propose a data-driven BFC metric to decide when to \textit{switch} from exploration to exploitation. We assume a fixed $\tau =\tau_0$ to begin and run PWLL as usual, without decaying $\tau_0$. At each iteration, PWLL acquires a new point according to the minimum norm acquisition function. Similar to the coreset method, we compute BFC between the new acquisition and the current set of all labeled points. This yields a data-driven metric indicating how explored the dataset is. 

When the metric is close to $-2$, this indicates the current acquisition is not ``close'' to any other points (as measured by BFC, see Definition \ref{def:bfc}), and PWLL should continue exploring. However, once exploration is complete, the acquisitions will get closer to one another, and BFC will increase. Assuming a kNN graph, the following argument provides a quantitative bound on when acquisitions are still far enough apart to continue exploring:
\begin{align} 
\text{Ric}(i,j) &= -2 + \frac{2}{d_i} + \frac{2}{d_j} \nonumber + \frac{2|\Delta(i,j)|}{\text{max}\{d_i,d_j\}} +\frac{|\Delta(i,j)|}{\text{min}\{d_i,d_j\}} + \frac{(\gamma_{\text{max}})^{-1} (|\Box^{i}(i,j)| + |\Box^{j}(i,j)|)}{\text{max}\{d_i,d_j\}} \nonumber \\
&= -2 + \frac{2}{d_i} + \frac{2}{d_j} \label{eq:pwll_bound_1} \\
&\leq -2 + \frac{4}{k}, \label{eq:pwll_bound_2} 
\end{align}
where \eqref{eq:pwll_bound_1} is because we assume $\Delta(i,j) = \Box^{i}(i,j) = \text{$\Box^{j}(i,j) = \emptyset$}$ when exploration is still in progress and \eqref{eq:pwll_bound_2} is due to the fact that $d_i \geq k$ for all $i$ in a kNN graph. Hence, an upper bound of $\text{Ric}(i,j)$ \textit{when $x_i$ and $x_j$ are not in the same neighborhood} (see Definition \ref{def:neighborhood}) is $-2 + \frac{4}{k}$. $\text{Ric}(i,j)$ exceeding this amount indicates that $x_i$ and $x_j$ are in the same neighborhood, and thus exploration is complete. We describe our methodology in Algorithm \ref{alg:pwll_curvature}.

\begin{algorithm}[t]
\caption{PWLL-$\tau$ with Curvature-based $\tau$ Schedule}\label{alg:pwll_curvature}
\begin{algorithmic}[1]

\Require PWLL-$\tau$ classifier with initial $\tau=\tau_0 >0$, minimum norm acquisition function $\mathcal{A}$, $k$ from $k$-NN search, number of AL iterations $n$.

\Ensure Labeled set of points $\mathcal{L}$.

\State Initialize $\mathcal{L} = \emptyset$

\For{$k$ = $1$ to $n$}

\State Acquire new point $x_i$ with $\mathcal{A}$

\State Compute $\text{Ric}(i,j)$ for each $x_j \in \mathcal{L}$\Comment{Definition \ref{def:bfc}}

\If{$\max_{j \in \mathcal{L}}\text{Ric}(i,j) > (-2 + \frac{4}{k})$} \Comment{Equation \ref{eq:pwll_bound_2}}

\State Update $\tau = 0$

\EndIf

\State Update $\mathcal{L} = \mathcal{L} \cup \{x_i\}$

\EndFor

\end{algorithmic}
\end{algorithm}

\subsection{Localized Graph Rewiring for Label Propagation} \label{sec:methods:rewiring}

While effective in some settings, single-scale graph constructions are sensitive to the choice of scale parameter and may fail to capture both local and global structure in heterogeneous data. In \textit{multiscale graph construction}, edge formation aggregates information across multiple neighborhood scales, enhancing representations of distinct clusters and density variations.
The multiscale structure can be used to more effectively regularize within well-connected regions of the graph - such as dense clusters - by penalizing higher-order variations of the labeling function. 
For example, in multiscale graph Laplacian regularization \cite{Merkurjev} and higher-order hypergraph learning \cite{weihs2025HOHL,weihs2025Hypergraphs}, powers of the Laplacian are applied locally depending on the graph’s structure, enabling adaptive smoothness control across different regions of the data.

Formally, we pick scale parameters $\varepsilon_1 > \dots > \varepsilon_q$ and construct the associated Laplacian matrices $\{L_{\varepsilon_k}\}_{k=1}^q$. Picking powers $p_1 \leq \dots \leq p_q$, we then use the regularizer \begin{align}
J(v) &= \sum_{k=1}^q \lambda_k v^\top L_{\varepsilon_k}^{p_k} v = v^\top \left[ \sum_{k=1}^q \lambda_k  L_{\varepsilon_k}^{p_k} \right] v =: v^\top L^{(q)} v \label{eq:HOHL}
\end{align}
where $\lambda_k$ are fixed positive weights.

\begin{figure}[ht]
    \centering
    \includegraphics[width=.4\linewidth]{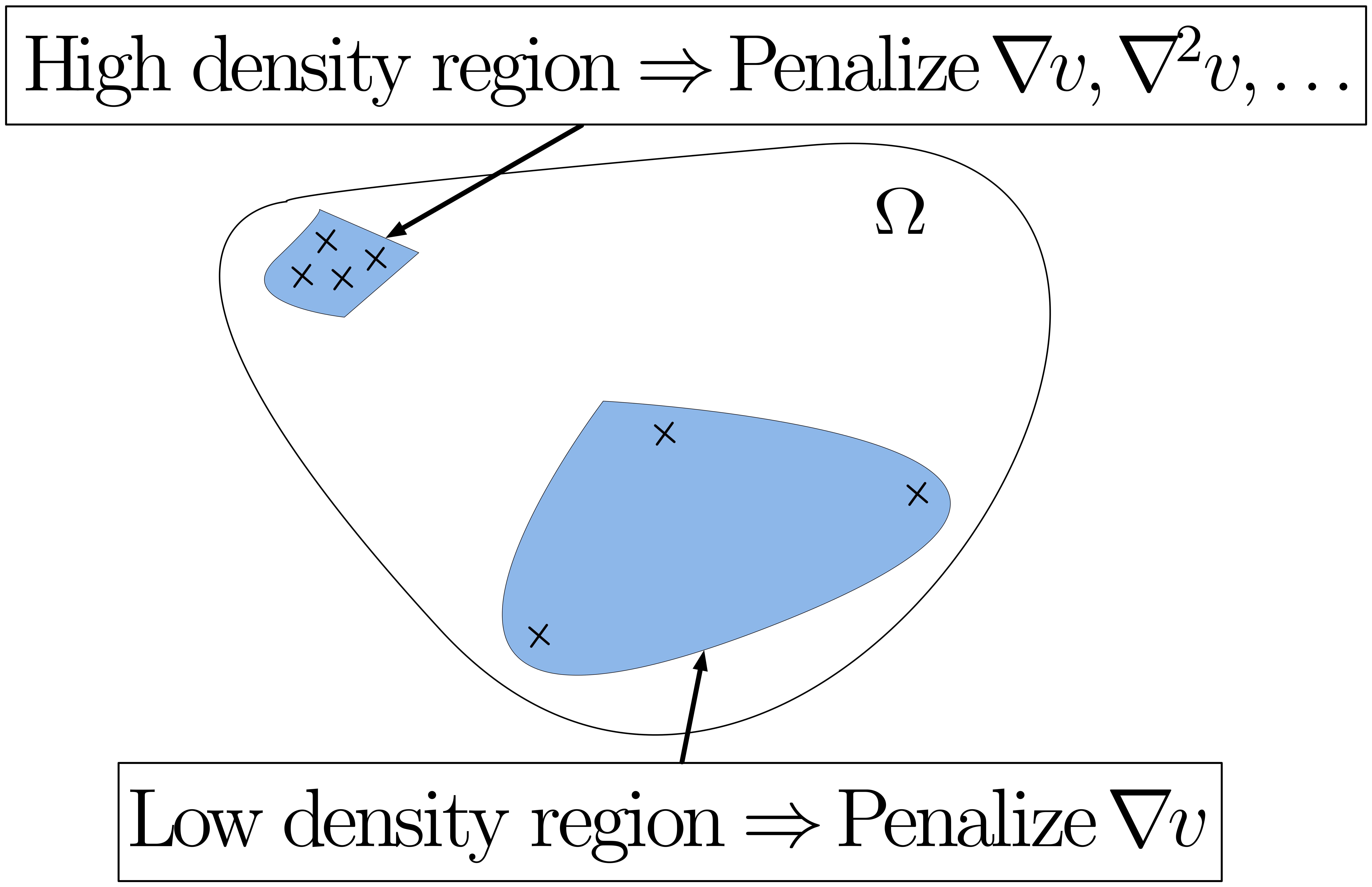}
    \caption{Higher-order smoothness is imposed on the labeling function $v$ in denser regions, while allowing greater flexibility in sparser areas \cite{weihs2025Hypergraphs}.}
    \label{fig:HDR}
\end{figure}

As $k$ increases, the associated scale $\varepsilon_k$ decreases, so the constructed graph retains only the strongest (most local) connections - emphasizing finer structural features and high-density neighborhoods. Under the standard graph homophily assumption \cite{ma2023homophilynecessitygraphneural}, these regions benefit from stronger regularization - encouraging smooth label variation across densely connected neighborhoods. This motivates using larger powers $p_k$ for the corresponding Laplacians $L_{\varepsilon_k}$. See Figure~\ref{fig:HDR} for a visual illustration of this intuition (e.g., with $p_k = k$).

The use of the multiscale regularizer $J(v) = v^\top L^{(q)} v$ in semi-supervised learning has been empirically validated in \cite{Merkurjev}. Its theoretical properties - including well-posedness, extensions to supervised learning, connections to hypergraph models, and applications beyond geometrically embedded data - have been further explored in \cite{weihs2025HOHL,weihs2025Hypergraphs}. Notably, it was shown that $L^{(q)}$ can itself be interpreted as the Laplacian of a derived graph, and its spectral characteristics have been analyzed in detail.

Graph-based active learning can be greatly improved by incorporating multiscale Laplacian regularization, i.e., replacing the classical graph Laplacian with $L^{(q)}$, as demonstrated in \cite{weihs2025HOHL}. However, computing $L^{(q)}$ incurs substantial computational cost: while the base Laplacian $L_\varepsilon$ is typically sparse, its powers $L_\varepsilon^k$ (for $k \geq 2$) grow increasingly dense.

Specifically, the entry \( (i,j) \) of \( L^k_\varepsilon \) is recursively given by
\(
(L^k_\varepsilon)_{ij} = \sum_{r=1}^n (L^{k-1}_\varepsilon)_{ir} \cdot (L_\varepsilon)_{rj},
\)
which expands explicitly to
\[
(L^k_\varepsilon)_{ij} = \sum_{i_1, i_2, \dots, i_{k-1}=1}^n (L_\varepsilon)_{i i_1} (L_\varepsilon)_{i_1 i_2} \cdots (L_\varepsilon)_{i_{k-1} j}.
\]
By construction, \( (L_\varepsilon)_{ij} \neq 0 \) if and only if \( w_{\varepsilon,ij} \neq 0 \), i.e., there exists an edge between nodes \( x_i \) and \( x_j \). Consequently, \( (L^k_\varepsilon)_{ij} \neq 0 \) if and only if there exists at least one path of length \( k \) between \( x_i \) and \( x_j \). As $k$ increases, the number of such paths - and thus nonzero entries - grows rapidly. This densification breaks the original sparsity pattern, resulting in significantly higher memory and computational costs. 

To circumvent this issue, we propose a localized approximation to multiscale regularization by rewiring the graph at only the labeled nodes, where higher-order smoothing is most beneficial. As discussed in Section \ref{sec:background:GLR}, reinforcing regularization of the labeling function near given points promotes propagation of their labels to neighboring nodes. Since the acquisition function targets high-utility samples, concentrating regularization around them enhances exploitation - particularly in homophilous graphs, where nearby nodes often share labels. Our approach is detailed in Algorithm \ref{alg:rewiring}.

We refer to our method as rewiring: at each iteration, the matrix $L$ in the regularizer $J(v)$ represents the Laplacian of an incrementally modified graph \cite{weihs2025HOHL}. Crucially, the updates to $L$ are additive and localized, enabling efficient incorporation of higher-order structure while avoiding the computation and densification of full Laplacian powers.

This approach hinges on the assumption that labeled nodes are particularly suitable targets for enhanced regularization, as they are \emph{purposefully} selected by the acquisition function $\mathcal{A}(x)$. To test this hypothesis, we also compare our method to a control strategy that applies the same localized regularization, but around randomly selected nodes at each step, rather than those chosen by $\mathcal{A}(x)$ (see Section~\ref{sec:results:rewiring}).

\begin{algorithm}[!t]
\caption{Localized Graph Rewiring for Label Propagation}\label{alg:rewiring}
\begin{algorithmic}[1]
\Require Dataset $\mathcal{X}$, labeling budget $B$, acquisition function $\mathcal{A}$, scale parameters $\varepsilon_1 > \dots > \varepsilon_q$, powers $p_1 \leq \dots \leq p_q$, positive coefficients $\{\lambda_k\}_{k=1}^q$.
\Ensure Labeled set $\mathcal{L}_B$ and corresponding classifier $u$
\State Compute weighted graph $G = (\{x_i\}_{i=1}^n, W_{\varepsilon_1})$ and matrix $L = L_{\varepsilon_1}^{p_1}$
\State Initialize coreset $\mathcal{L}_0 \subset \mathcal{X}$, set $\mathcal{U}_0 = \mathcal{X} \setminus \mathcal{L}_0$
\For{$i = 0$ to $B - 1$}
    \State Train classifier $u_i$ using labeled set $\mathcal{L}_i$ through Laplace learning \Comment{Section \ref{sec:background:GLR}}
    \State Select query point: \(
        x_{\mathrm{acq}} = \argmax_{x \in \mathcal{U}_i} \mathcal{A}(x)
    \)
    \State Update labeled and unlabeled sets: \(
        \mathcal{L}_{i+1} = \mathcal{L}_i \cup \{x_{\mathrm{acq}}\}, \quad
        \mathcal{U}_{i+1} = \mathcal{U}_i \setminus \{x_{\mathrm{acq}}\}
    \)
    \State Update Laplacian: \[
        L \gets L + \sum_{k=2}^q \lambda_k (L_{\varepsilon_k}^{\{x_{\mathrm{acq}}\}})^{p_k} 
    \]
\EndFor
\State Train final classifier $u$ using labeled set $\mathcal{L}_B$
\end{algorithmic}
\end{algorithm}

\section{Numerical Experiments} \label{sec:results}

\subsection{Graph Construction and Active Learning Setup}

For all experiments, following the discussion in Section~\ref{sec:background:graph_construct}, we construct kNN graphs\footnote{We use Annoy for approximate nearest neighbor searches \url{https://github.com/spotify/annoy}.}, replacing the scale parameters $\varepsilon_k$ with a decreasing sequence of neighborhood sizes $k_1 > \dots > k_q$ ($q=1$ except for the multiscale rewiring experiments in Section \ref{sec:results:rewiring}). This ensures that our weight matrices are sparse, lowering memory and compute overhead. Because GBSSL is a transductive paradigm, we embed and build the graph over the entire dataset, irrespective of built-in training-testing splits.

Edges are weighted with a gaussian kernel
\[w_{ij} = \text{exp}\Bigl( \frac{-4 \Vert x_i - x_j \Vert^2}{d_k(x_i)^2} \Bigr), \]
where $d_k(x_i)$ is the distance from $x_i$ to its $k$th nearest neighbor, which ensures edge weights adapt to both sparse and dense regions of the feature space. The norm $\Vert \cdot \Vert$ is cosine similarity (angular metric). Importantly, weights are computed only for the $k$ nearest neighbors identified via the kNN search; all other entries in $W$ are set to zero.

Sections~\ref{sec:basic_al_results} and \ref{sec:results:sequential_vs_batch} present coreset and AL results on image datasets, comparing our proposed CC with other coreset methods on downstream sequential and batch AL, respectively. 
In Section \ref{sec:results:PWLL}, we report results on image datasets for our BFC-based metric that provides PWLL-$\tau$ a data-driven signal to inform the transition from exploration to exploitation. 
Finally, in Section~\ref{sec:results:rewiring}, we adopt the localized rewiring strategy described in Algorithm~\ref{alg:rewiring}. 
Across all experiments - except PWLL-$\tau$ in Section \ref{sec:results:PWLL} - we use uncertainty sampling as the acquisition function (see Section~\ref{sec:background:AL}). Throughout, plots display mean accuracy (solid line) with one standard deviation (shaded region) across trials.

\subsection{Coreset and Sequential Active Learning on Image Benchmarks}\label{sec:basic_al_results}

\begin{figure}[!ht]
    \centering
    \begin{subfigure}[t]{0.32\linewidth} 
    \includegraphics[width=\linewidth]{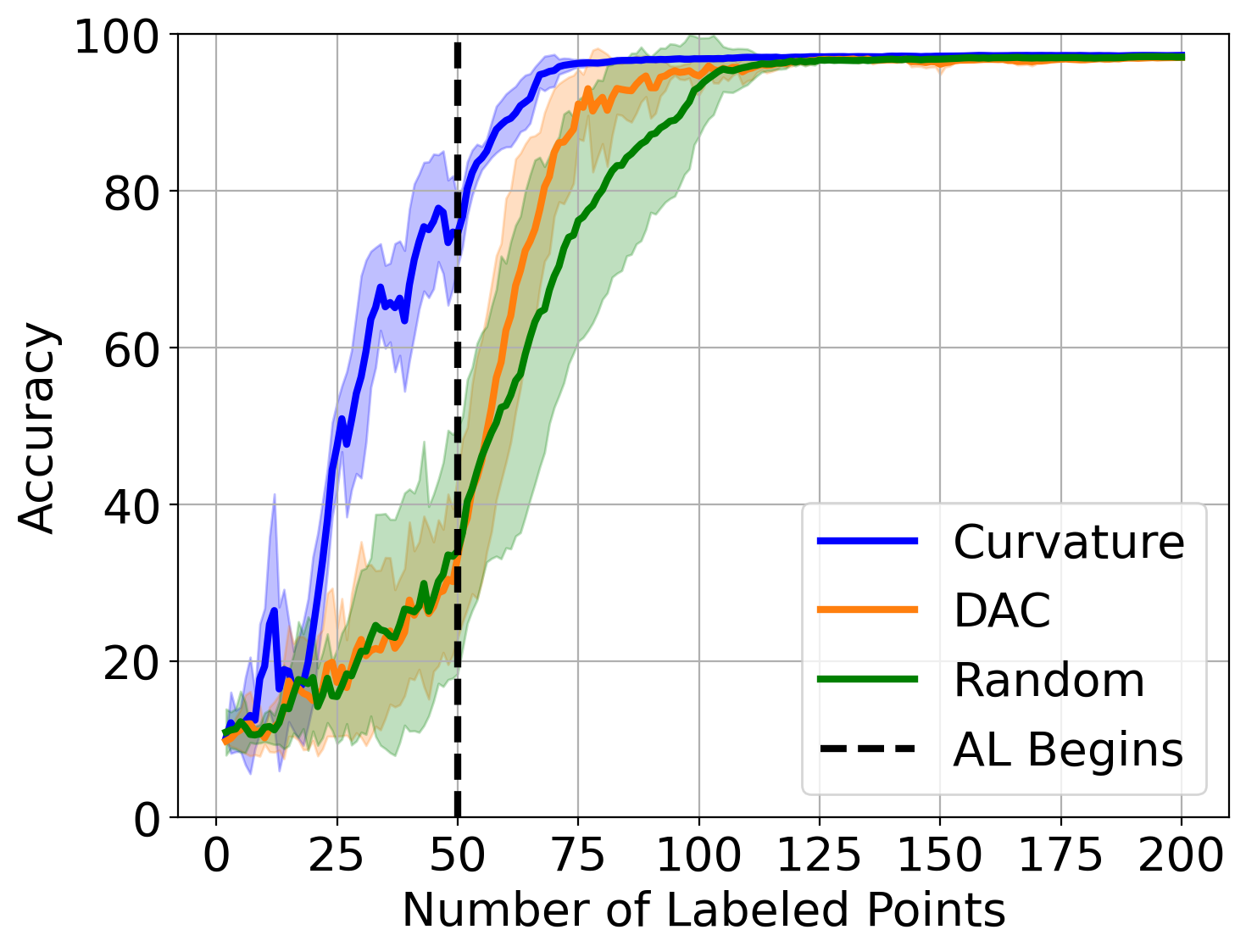}
    \caption{MNIST, AL begins at 50 labels.}
    \end{subfigure}
    \hfill
    \begin{subfigure}[t]{.32\linewidth}
    \includegraphics[width=\linewidth]{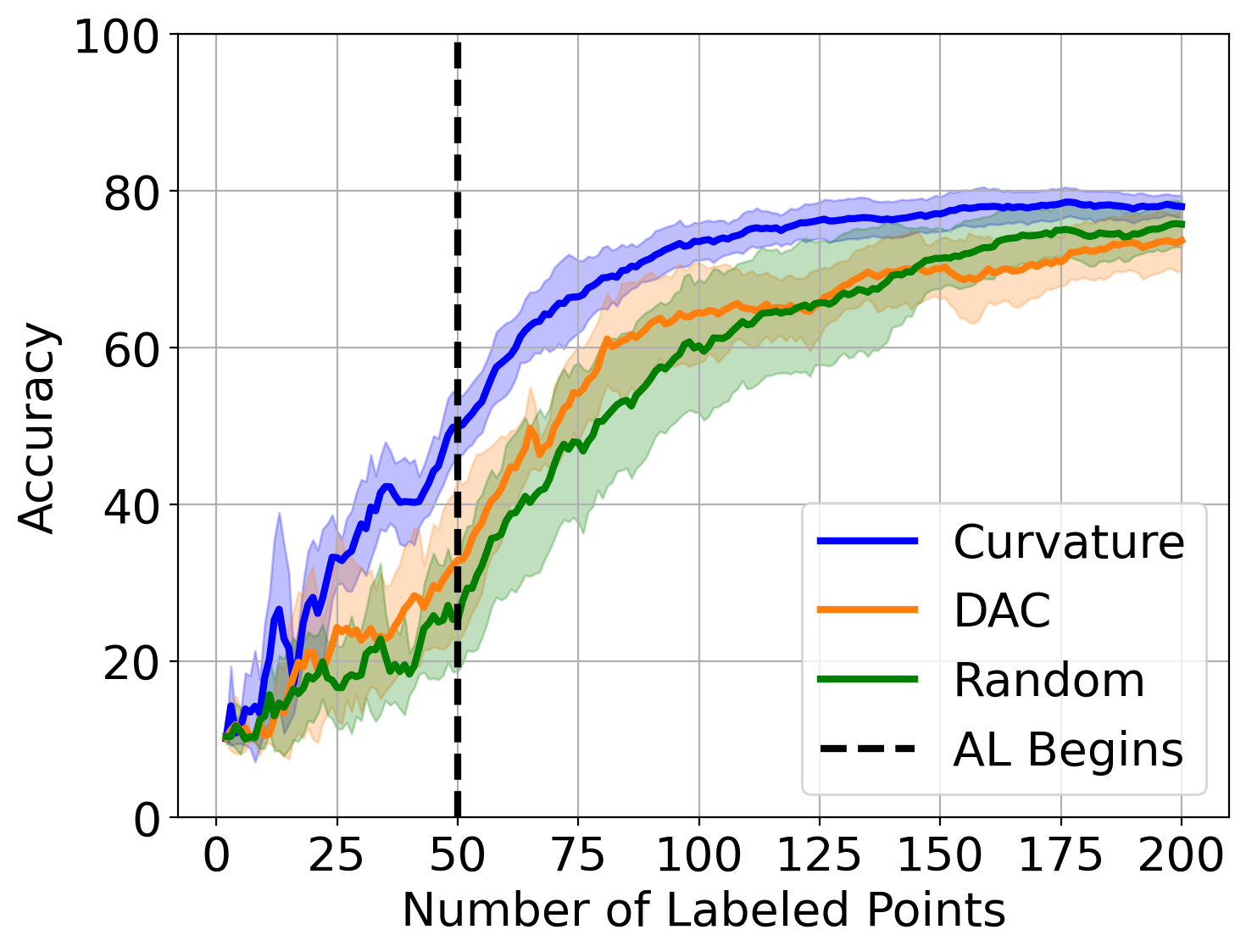}
    \caption{FashionMNIST, AL begins at 50 labels.}
    \end{subfigure}
    \hfill
    \begin{subfigure}[t]{.32\linewidth} 
    \includegraphics[width=\linewidth]{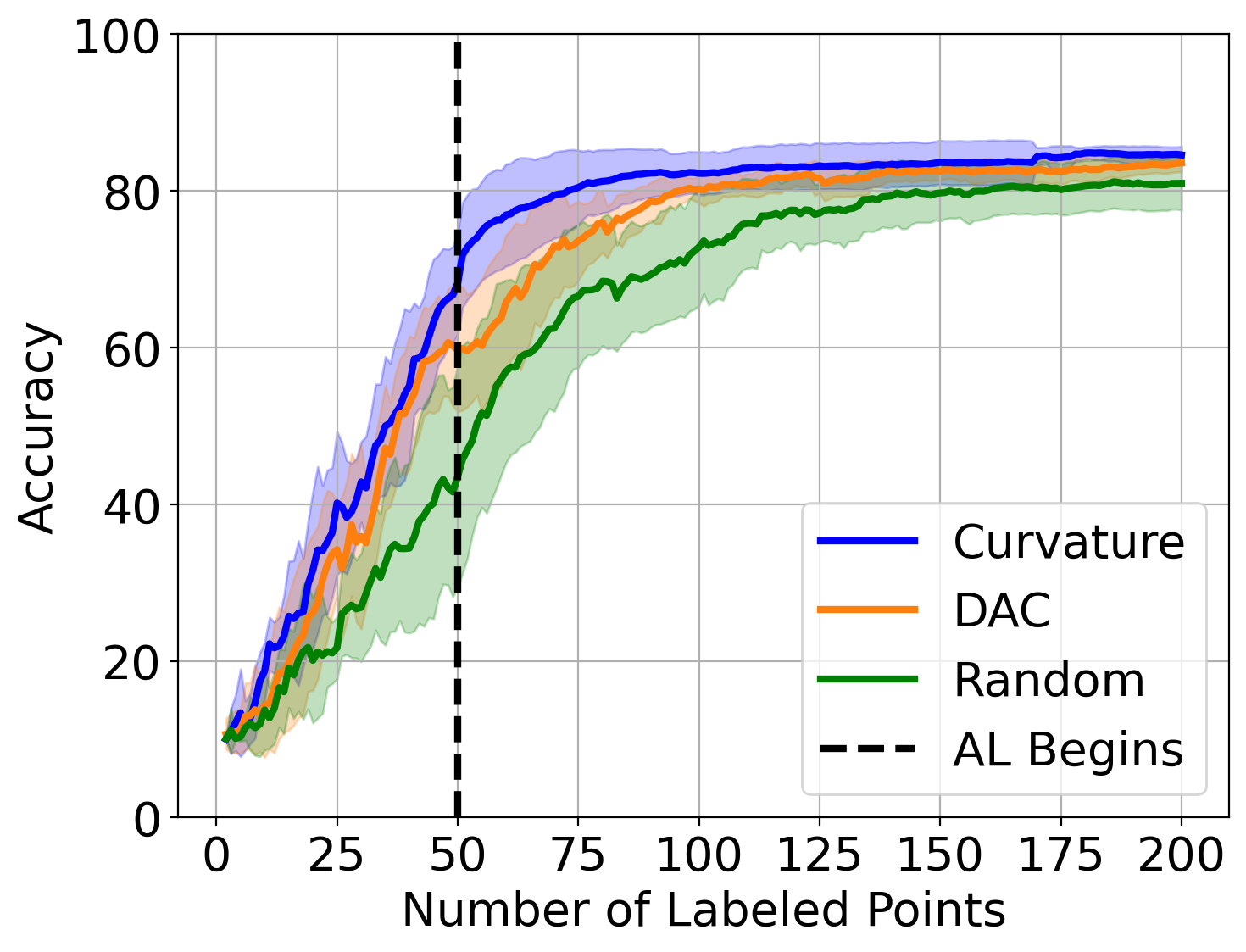}
    \caption{CIFAR-10, AL begins at 50 labels.}
    \end{subfigure}
    \hfill
    \begin{subfigure}[t]{0.32\linewidth}  
    \includegraphics[width=\linewidth]{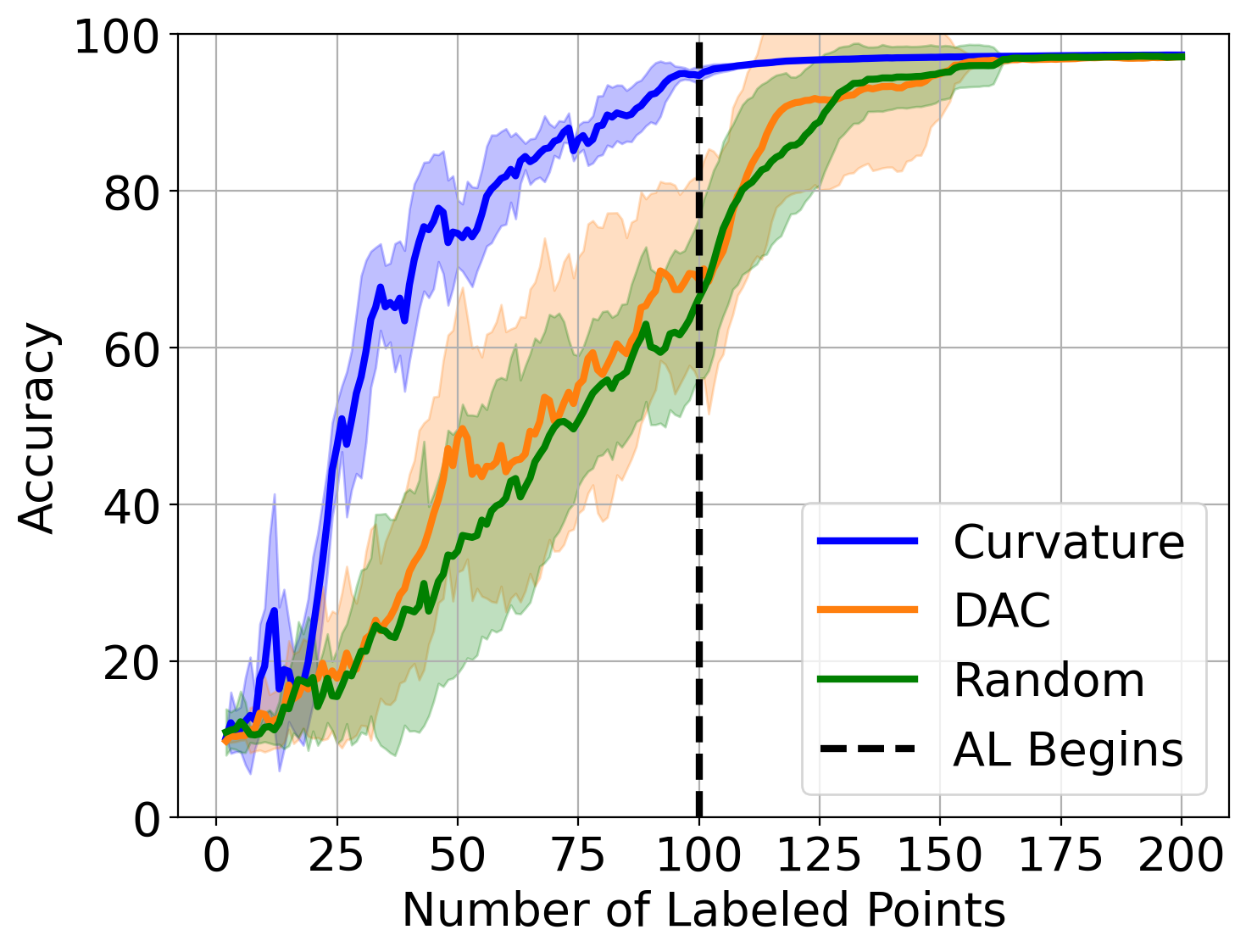}
    \caption{MNIST, AL begins at 100 labels.}
    \end{subfigure}
    \hfill
    \begin{subfigure}[t]{0.32\linewidth}  
    \includegraphics[width=\linewidth]{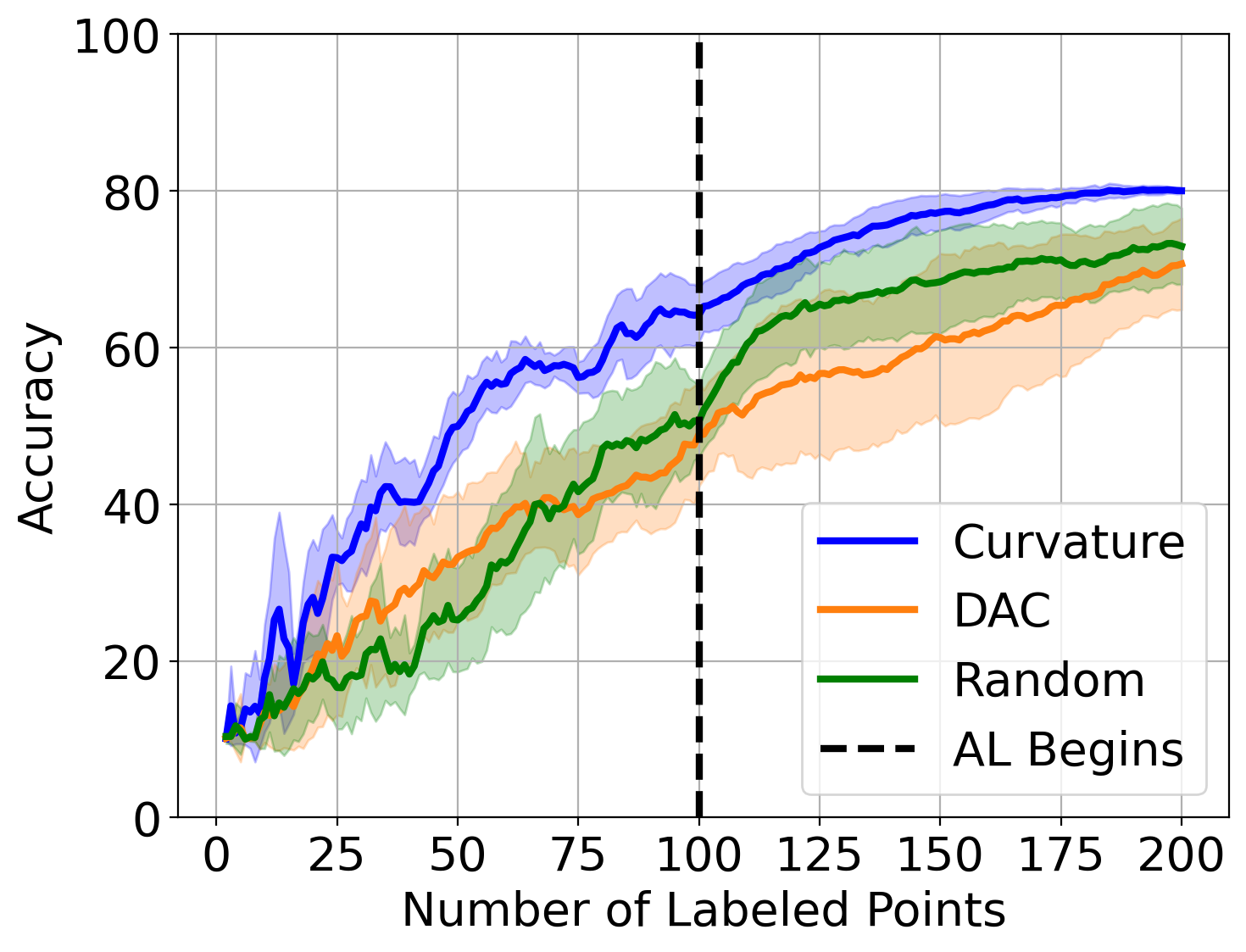}
    \caption{FashionMNIST, AL begins at 100 labels.}
    \end{subfigure}
    \hfill
    \begin{subfigure}[t]{0.32\linewidth}  
    \includegraphics[width=\linewidth]{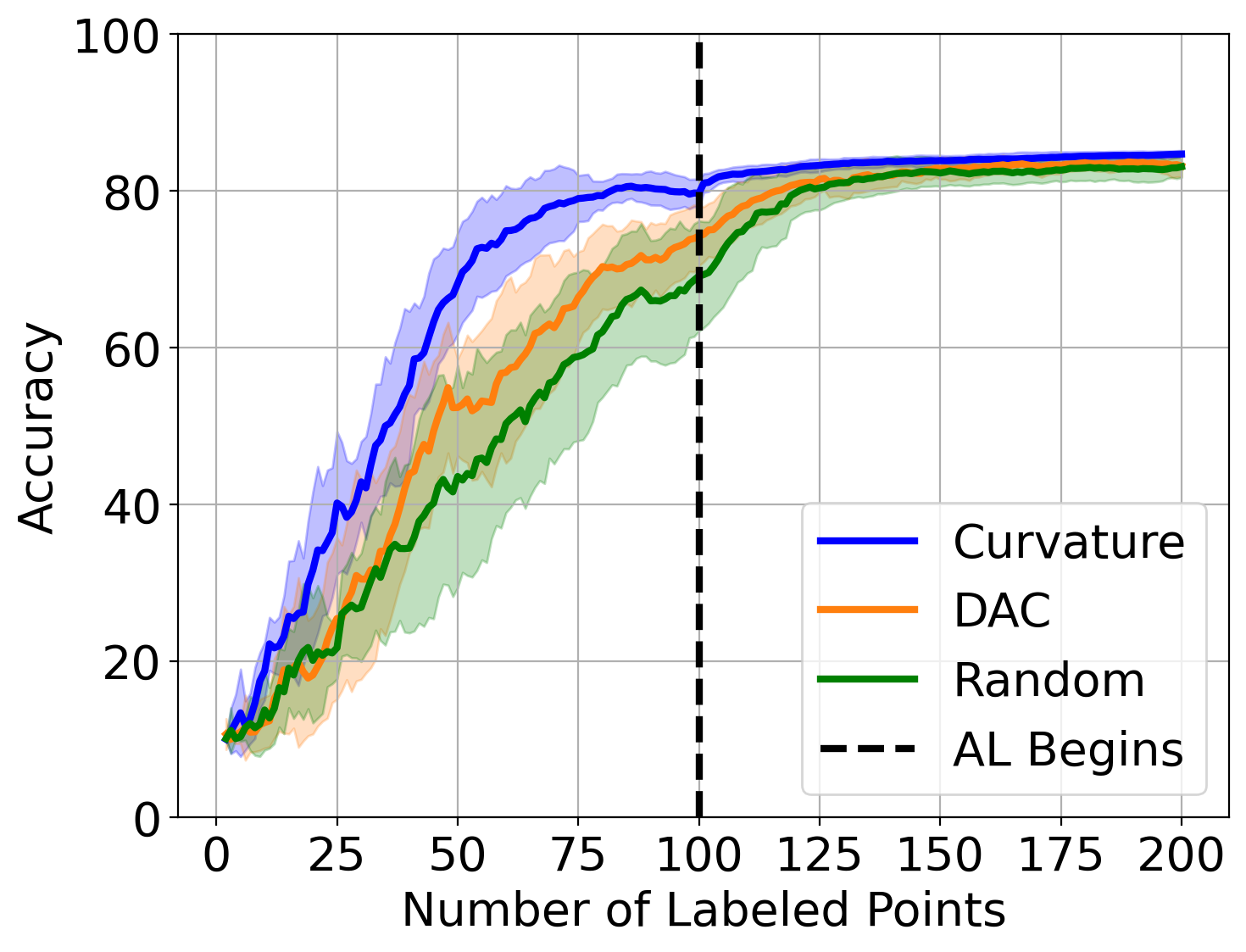}
    \caption{CIFAR-10, AL begins at 100 labels.}
    \end{subfigure}
    \caption{Coreset and AL results for Curvature, DAC, and Random on several benchmarks. The top and bottom rows present results when AL begins at 50 and 100 labels, respectively. For reference, 50 labeled points is approximately 0.07\% of the MNIST and FashionMNIST datasets and 0.08\% of CIFAR-10. The solid line indicates the mean and shaded region indicates one standard deviation over 10 trials.}
    \label{fig:coreset_100}
\end{figure}

\begin{figure*}[!ht]
    \centering
    \begin{subfigure}[t]{0.32\linewidth}  
    \includegraphics[width=\linewidth]{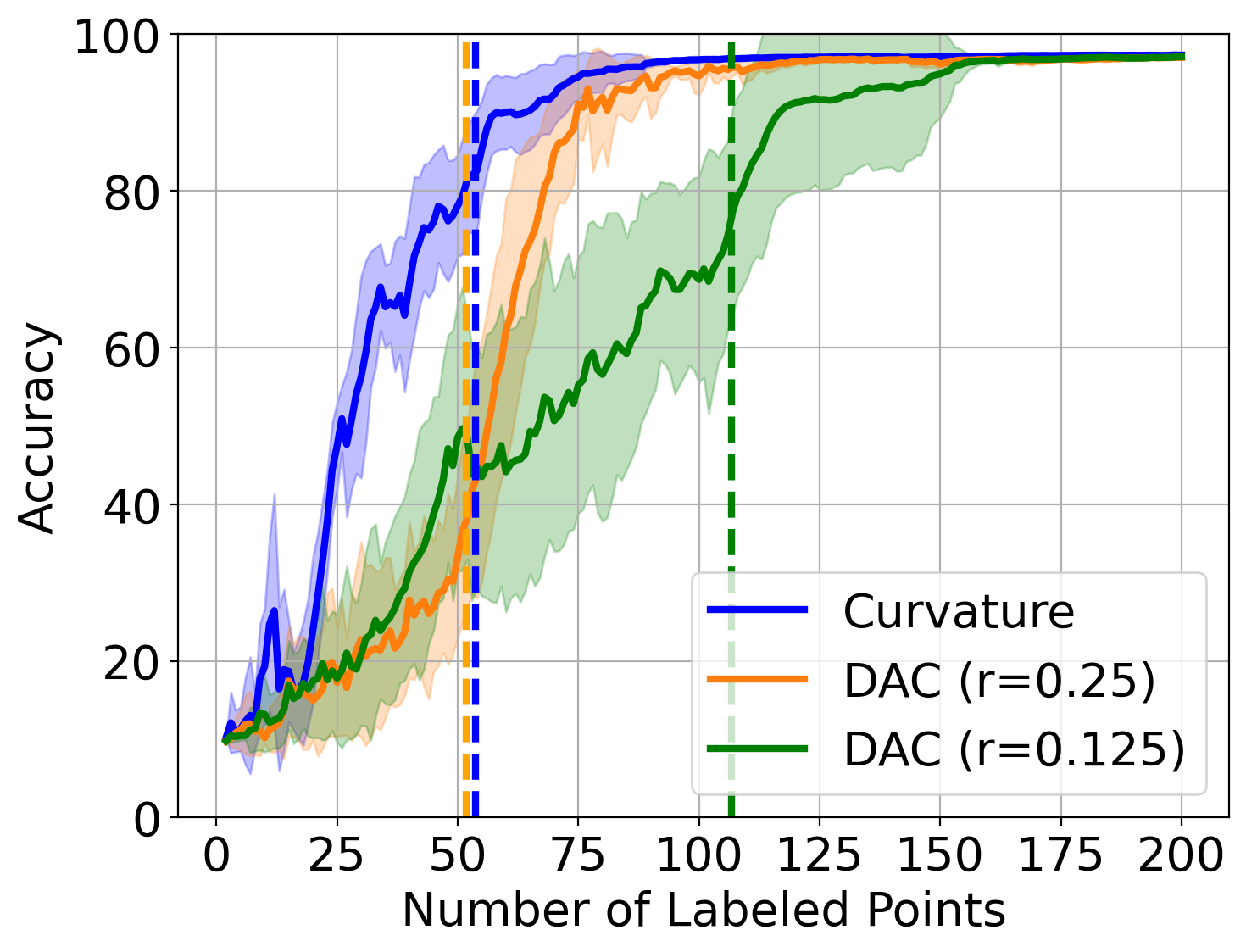}
    \caption{MNIST}
    \end{subfigure}
    \hfill
    \begin{subfigure}[t]{0.32\linewidth}  
    \includegraphics[width=\linewidth]{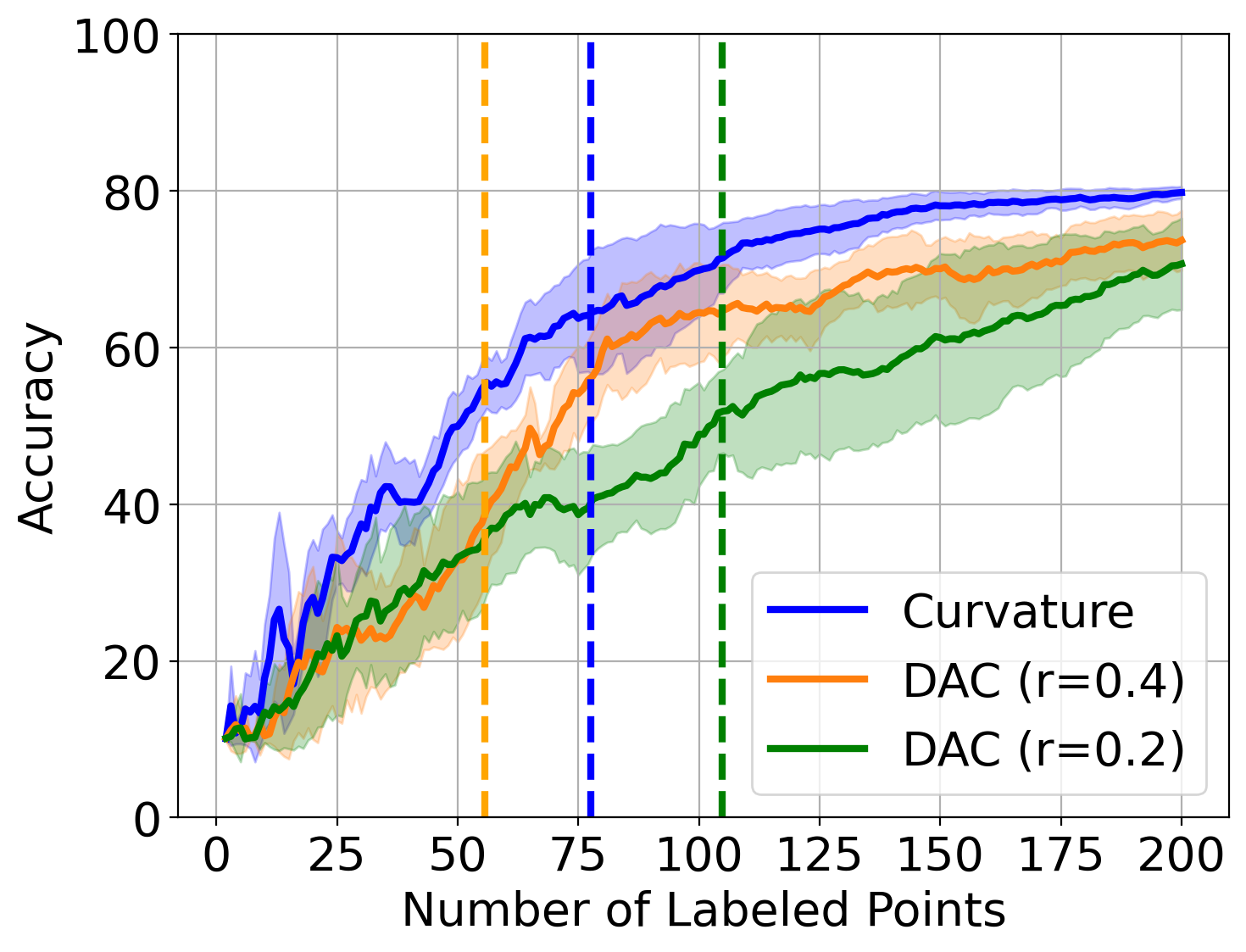}
    \caption{FashionMNIST}
    \end{subfigure}
    \hfill
    \begin{subfigure}[t]{0.32\linewidth}  
    \includegraphics[width=\linewidth]{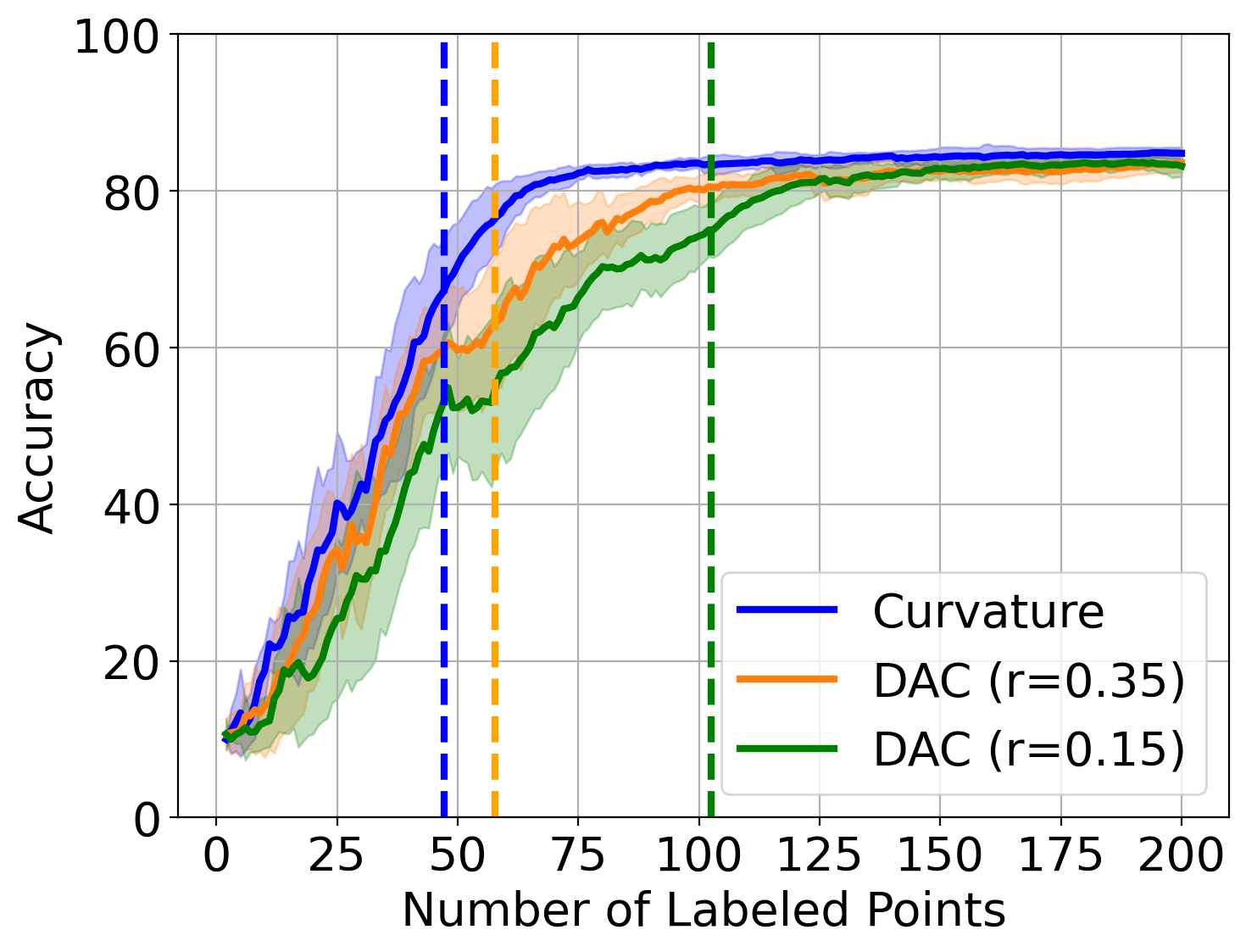}
    \caption{CIFAR-10}
    \end{subfigure}
    \caption{Coreset and AL accuracy comparison between our method and DAC (with different radii), where we use each method's stopping condition. The like-colored dashed line indicates where the stopping condition is triggered (and AL begins) for each method. 
    }
    \label{fig:coreset_stopping}
\end{figure*}

We present results on the MNIST \cite{mnist}, FashionMNIST \cite{xiao2017fashion}, and CIFAR-10 \cite{cifar} datasets. 
Before constructing the graphs, we first embed the data into a latent space using unsupervised neural networks, which provides a better setting to compute similarities and ensure a meaningful graph \cite{brown2024gll,miller_graph-based_2022}.
For MNIST and FashionMNIST, we utilize VAE \cite{vae} embeddings, and similarly we use SimCLR \cite{simclr} embeddings of the CIFAR-10 data, all provided by the \texttt{graphlearning} Python package \cite{graphlearning}. 
We construct the graph with $k_1=25$, and use Laplace learning as the classifier (see Section~\ref{sec:background:GLR}). We run 10 trials of each experiment.
Our main results are shown in Figures \ref{fig:coreset_100} and \ref{fig:coreset_stopping}, and Table \ref{tab:coresets}. We discuss them further here.

\begin{remark}[Presenting a Fair Comparison to DAC]\label{remark:DAC_R}
DAC requires the user to specify a radius $R$ that defines the distance from the current coreset points to the set of candidates for the next coreset point. The algorithm terminates once every coreset point is within at least $R$ of every other point. Hence, the choice of $R$ has a large impact on the number of points in the coreset, and thus also the accuracy. This leads to two issues: (1) since the algorithm is stochastic - randomly choosing a point from the set of candidate points at each iteration - it is impossible to ensure that the coreset will finish with a certain number of points, and (2) manually ending DAC early before it has reached its stopping condition could lead to catastrophic results; one could terminate DAC before a large portion of the dataset has been ``seen'' according to the choice of $R$. 
\end{remark}

Remark \ref{remark:DAC_R} highlights strengths of our CC method: it does not require any user-inputted parameter such as $R$ - which has a significant impact on the size of the coreset and its accuracy - and greedily chooses the best point at each iteration. However, this presents some difficulty in reporting completely fair comparisons between our method and DAC at a given label rate, such as $100$ labels in Figure \ref{fig:coreset_100}. As we have established, issue (1) may cause DAC to end with fewer than $100$ points, and truncating a coreset that exceeds $100$ labeled points could significantly harm results per issue (2). 

To present a fair comparison, our experiments with fixed-size coresets (Figure \ref{fig:coreset_100}) use the following procedure: we fix a coreset label budget (50 or 100), and report DAC results where we pick an $R$ that leads to approximately that many points. When the coreset ends, we start active learning (uncertainty sampling) and report results up to 200 total labels.
For CC and Random, we always stop the coreset at 50 (Figure \ref{fig:coreset_100}, top) or 100 (Figure \ref{fig:coreset_100}, bottom)  labeled points, since these do not have the same behavior as DAC. For DAC, we use $R=0.25$, $0.4$, and $.35$ for MNIST, FashionMNIST, and CIFAR-10, respectively to acquire approximately 50 points, and half those values for 100 points. \ref{sec:app:dac_coreset_sizes} lists statistics on the coreset sizes obtained by DAC across each of the experiments, illustrating its high variability.
For CC, we use a reduction parameter of $r=100$ for MNIST and FashionMNIST and $r=50$ for CIFAR-10.

\begin{table*}[t]
\centering
\renewcommand{\arraystretch}{1.2}
\setlength{\tabcolsep}{6pt}
\begin{tabular}{|l|cc|cc|cc|}
\hline
\textbf{Method} &
\multicolumn{2}{c|}{\textbf{MNIST}} &
\multicolumn{2}{c|}{\textbf{FashionMNIST}} &
\multicolumn{2}{c|}{\textbf{CIFAR-10}} \\
\hline
 & Acc. (\%) & Time (s) & Acc. (\%) & Time (s) & Acc. (\%) & Time (s) \\
\hline
\textbf{CC} & 94.7 & 39.6 & 64.1 & 38.1 & 79.7 & 22.1 \\
DAC & 68.7 & 62.1 & 48.9 & 76.1 & 74.2 & 46.2\\
Random & 66.3 & 0.0 & 50.5 & 0.0 & 69.1 & 0.0 \\
\hline
\end{tabular}
\caption{Comparison of accuracy and efficiency for coreset methods after 100 labels, averaged over 10 trials. CC is consistently faster and performs better than DAC and Random. 100 labels is a 0.14\% label rate for MNIST and FashionMNIST, and a 0.16\% label rate for CIFAR-10.}
\label{tab:coresets}
\end{table*}

Figure \ref{fig:coreset_100} shows that our novel CC method significantly outperforms DAC and Random across all three datasets, and leads to faster convergence on downstream AL. For example, our method combined with AL achieves 80\% accuracy on the FashionMNIST dataset with only 200 labels, corresponding to a 0.2\% label rate. This is an over 10\% absolute improvement over the other methods. On the other datasets, we see convergence of all three methods to approximately the same accuracy with AL, but CC does so with significantly fewer iterations (labels). We summarize our coreset results in Table~\ref{tab:coresets}, which includes timing details demonstrating that CC is faster to run than DAC, in addition to being more accurate.

Since the stopping condition of DAC is the crux of the issue outlined in Remark \ref{remark:DAC_R}, we also present results where AL starts \textit{after the coreset method's stopping condition is reached} in Figure \ref{fig:coreset_stopping}. Although this means AL may start at different iterations for different methods, it allows for a clear comparison of both methods (and their stopping conditions). Specifically, Figure \ref{fig:coreset_stopping} compares CC with DAC (at different values of $R$) where the coreset method ends according to its own stopping condition. A good stopping condition should terminate (and hence switch to AL) early enough to be label-efficient but not so early as to leave a portion of the graph unexplored. Figure \ref{fig:coreset_stopping} shows that - regardless of whether our method's stopping condition triggers before, during, or after DAC's - our method is more accurate before and during AL across datasets.

\subsection{Sequential vs. Batch Active Learning}\label{sec:results:sequential_vs_batch} 

\begin{figure*}[!ht]
    \centering
    \begin{subfigure}[t]{.32\linewidth} 
    \includegraphics[width=\linewidth]{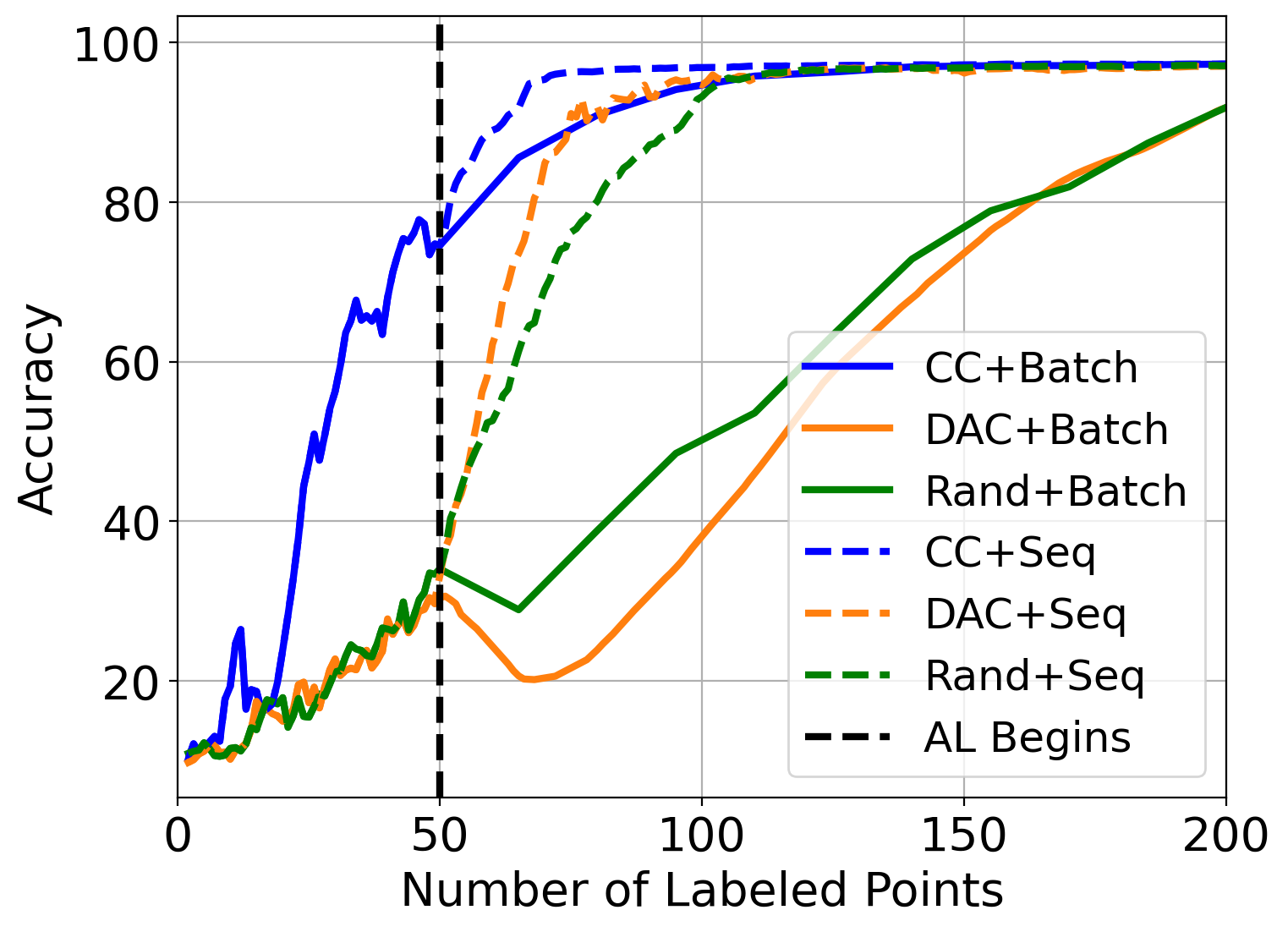}
    \caption{MNIST, AL begins at 50 labels.}
    \end{subfigure}
    \hfill
    \begin{subfigure}[t]{.32\linewidth}
    \includegraphics[width=\linewidth]{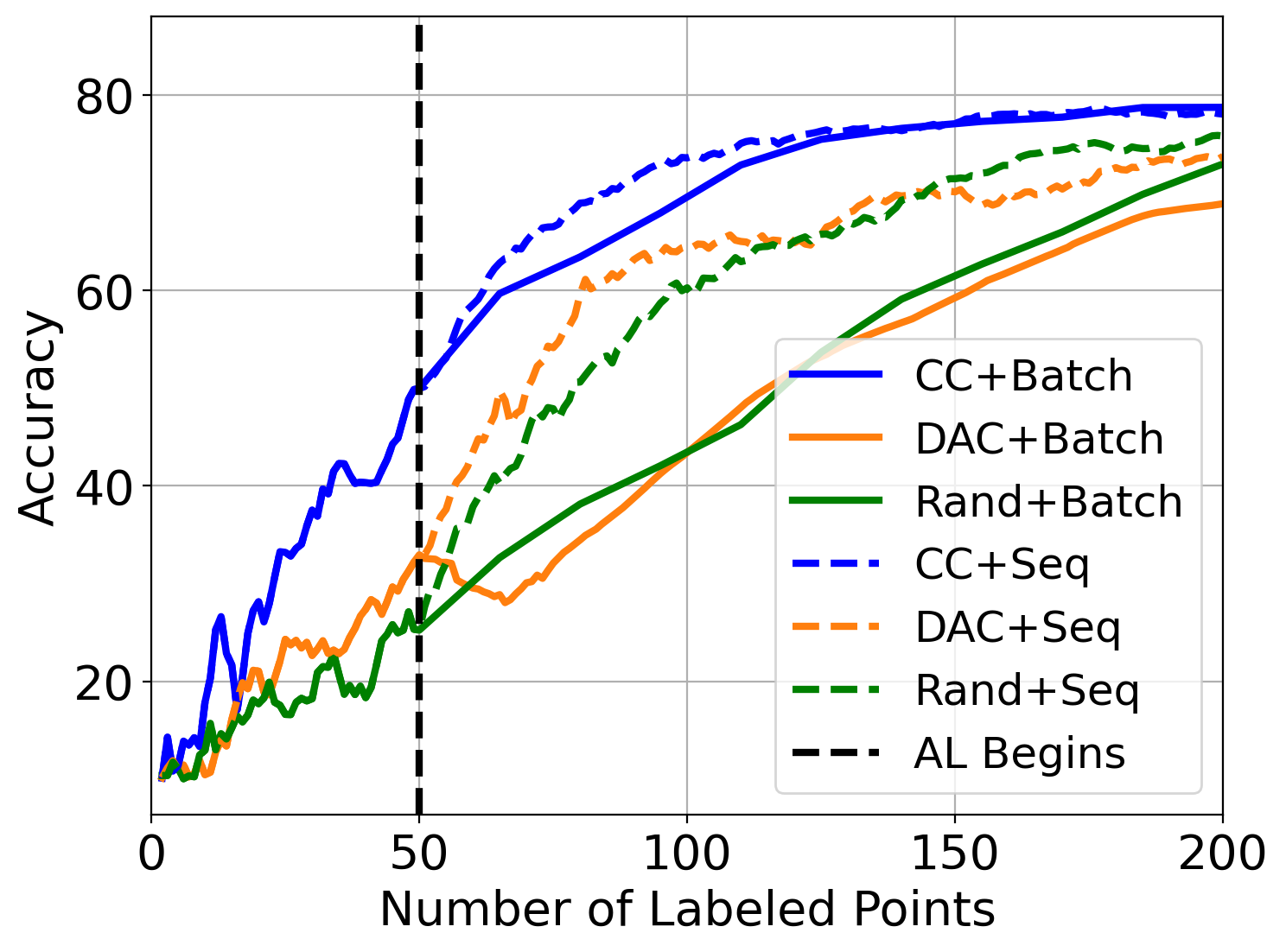}
    \caption{FashionMNIST, AL begins at 50 labels.}
    \end{subfigure}
    \hfill
    \begin{subfigure}[t]{.32\linewidth} 
    \includegraphics[width=\linewidth]{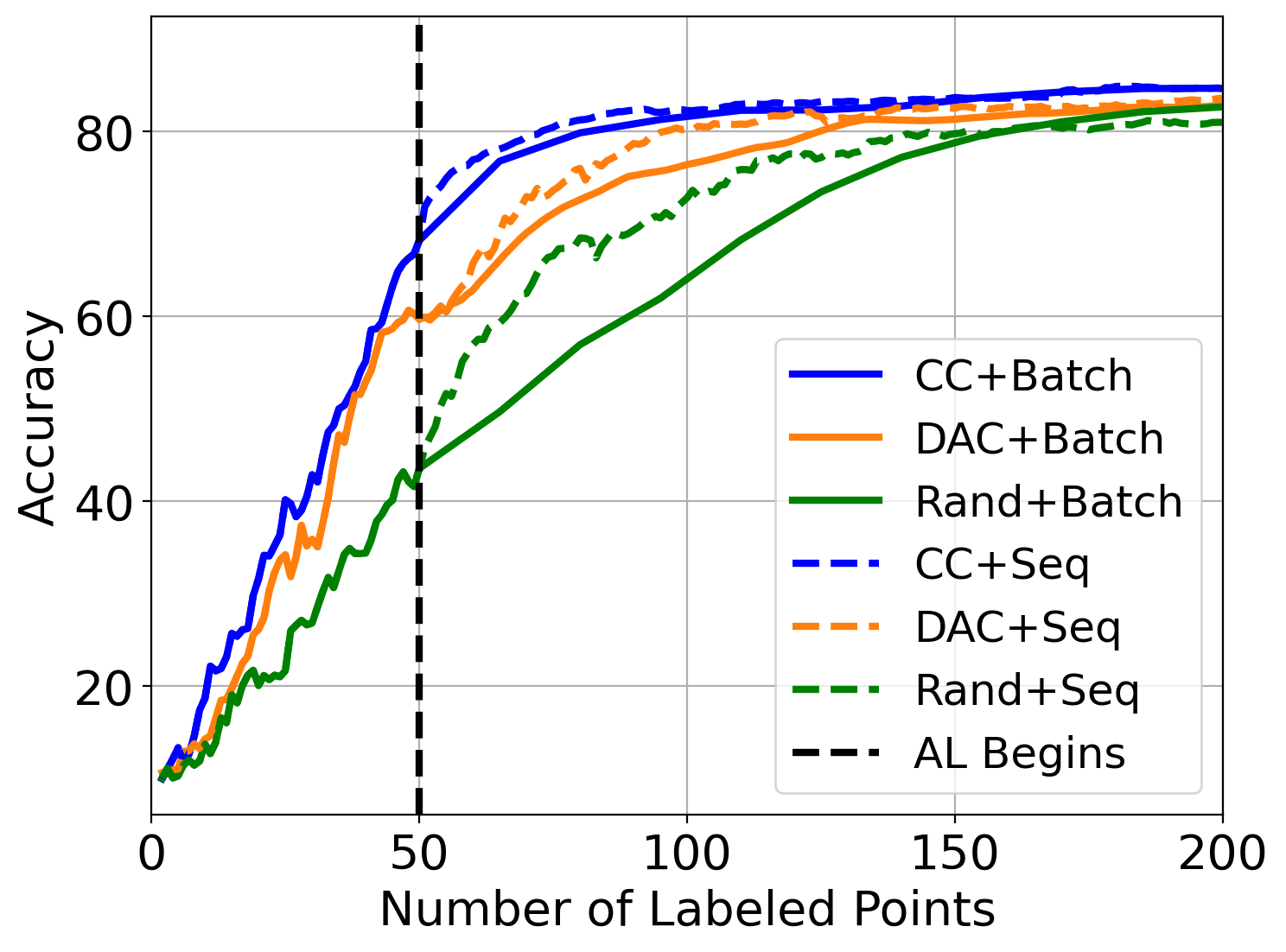}
    \caption{CIFAR-10, AL begins at 50 labels.}
    \end{subfigure}
    \hfill
    \begin{subfigure}[t]{0.32\linewidth}  
    \includegraphics[width=\linewidth]{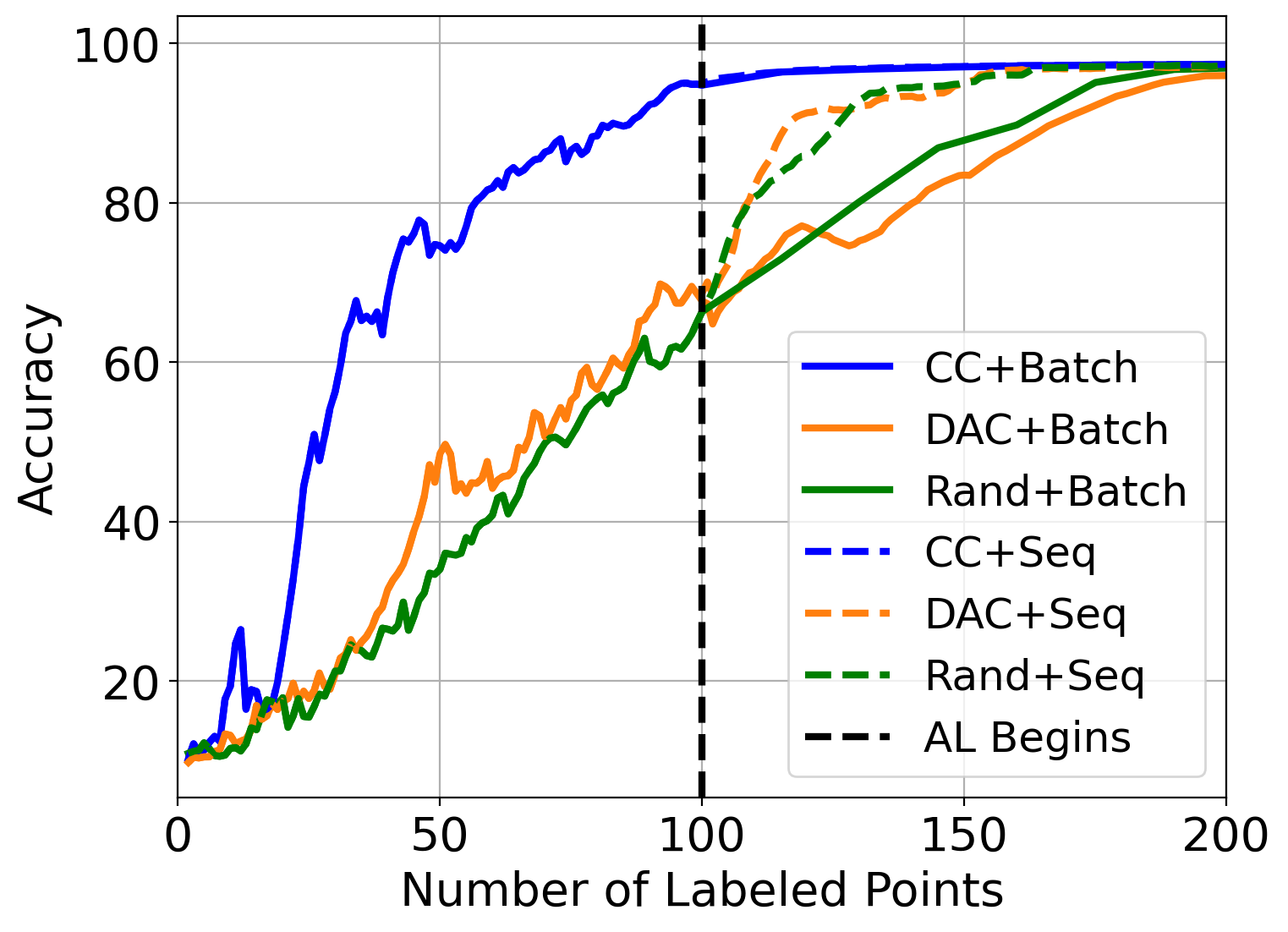}
    \caption{MNIST, AL begins at 100 labels.}
    \end{subfigure}
    \hfill
    \begin{subfigure}[t]{0.32\linewidth}  
    \includegraphics[width=\linewidth]{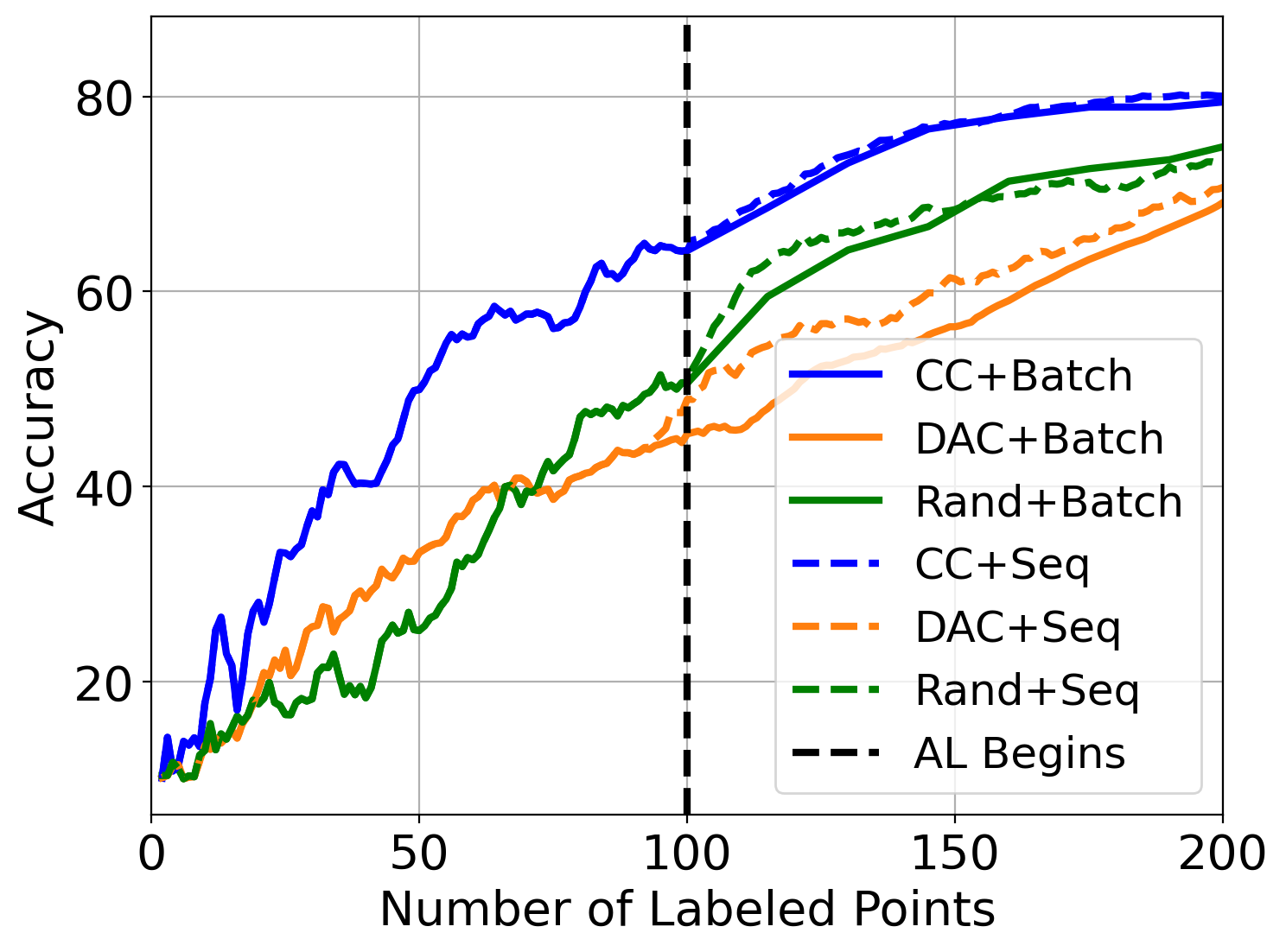}
    \caption{FashionMNIST, AL begins at 100 labels.}
    \end{subfigure}
    \hfill
    \begin{subfigure}[t]{0.32\linewidth}  
    \includegraphics[width=\linewidth]{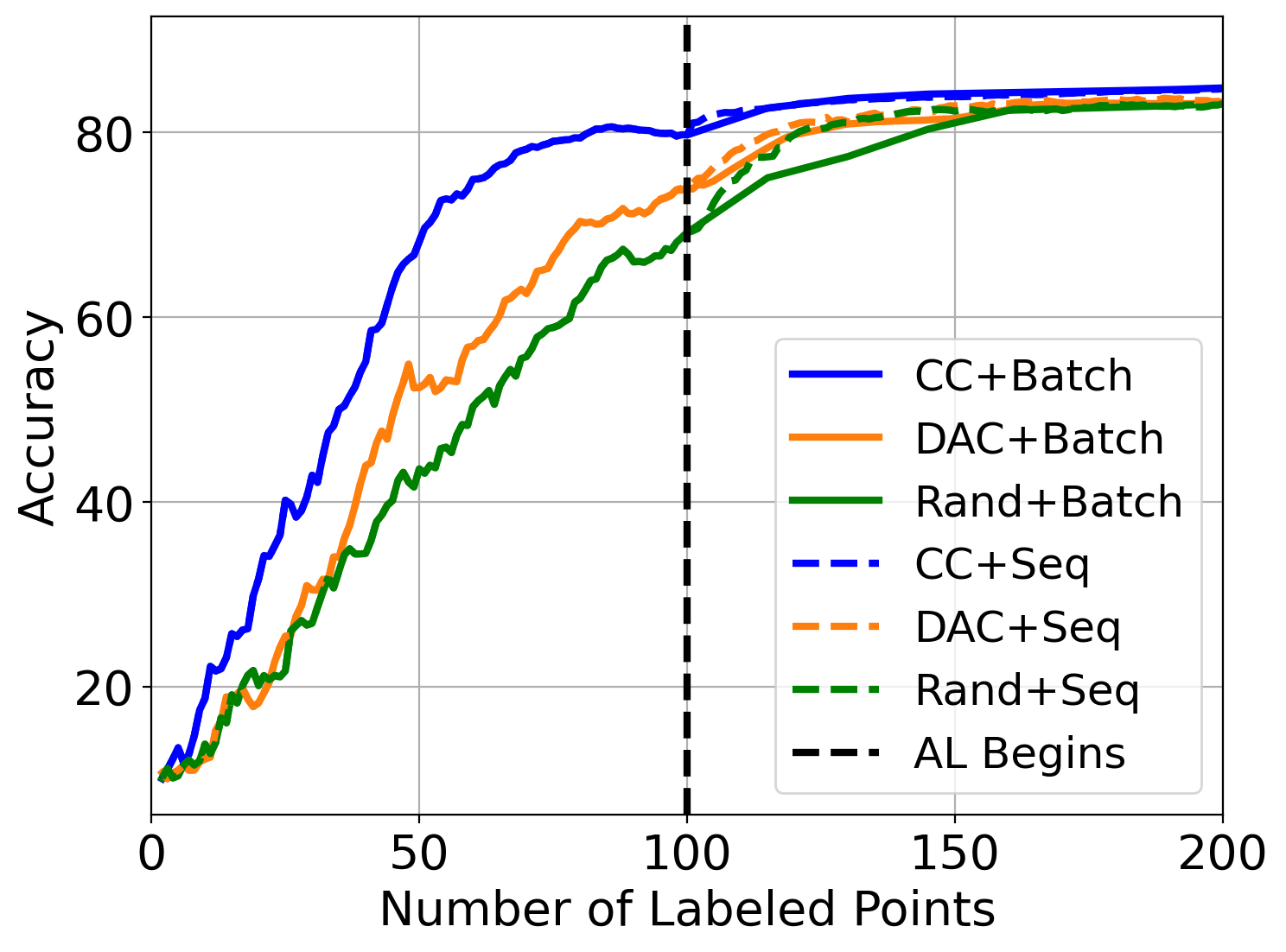}
    \caption{CIFAR-10, AL begins at 100 labels.}
    \end{subfigure}
    \caption{Coreset and AL results for Curvature, DAC, and Random on several image benchmarks. The top and bottom rows present results when AL begins at 50 and 100 labels, respectively. We compare downstream AL performance under batch (solid line) and sequential (dashed line) uncertainty sampling AL. We report the mean over 10 trials. We exclude standard deviation shading for clarity. 
    }
    \label{fig:batch_vs_seq_image}
\end{figure*}

Another empirical criterion to evaluate a coreset method is to compare the performance of downstream sequential and batch AL with the same initial coreset. Batch AL offers significant efficiency savings at the cost of accuracy compared to sequential AL, but a carefully chosen coreset can minimize the difference \cite{chapman_novel_2023}. In a sense, exploitative batch AL requires even stronger coreset exploration than in the sequential case; the underlying classifier must have sufficient information to determine not only one useful point to query, but several.

Figure \ref{fig:batch_vs_seq_image} compares CC, DAC, and Random coresets on MNIST, FashionMNIST, and CIFAR-10, reporting both sequential and batch AL accuracy with Laplace learning over 10 trials\footnote{Note that the sequential AL results in Figure \ref{fig:batch_vs_seq_image} are the same as in Figure \ref{fig:coreset_100}, but we reproduce them for a clear comparison to batch AL.}. We use LocalMax as our batch policy \cite{chapman_novel_2023} with a batch size of 15. Not only does CC provide the best coreset for both AL policies, but the decrease in accuracy between sequential and batch AL is minimal. Conversely, batch AL after DAC and Random is significantly worse than sequential AL - in some cases \textit{decreasing} in accuracy when batch AL begins. This implies that these coreset methods are not selecting effective initial labels for querying in batches; there is enough information in the coreset to choose good sequential queries (and update the classifier more frequently), but not enough to choose good batches. Indeed, AL behavior with DAC coresets is much closer to Random than it is to CC, implying a topologically-aware, greedy strategy is better for ensuring meaningful exploration.

\subsection{Application to PWLL} \label{sec:results:PWLL}

\begin{figure*}[!ht]
    \centering
    \begin{subfigure}[t]{0.45\linewidth}  
    \includegraphics[width=\linewidth]{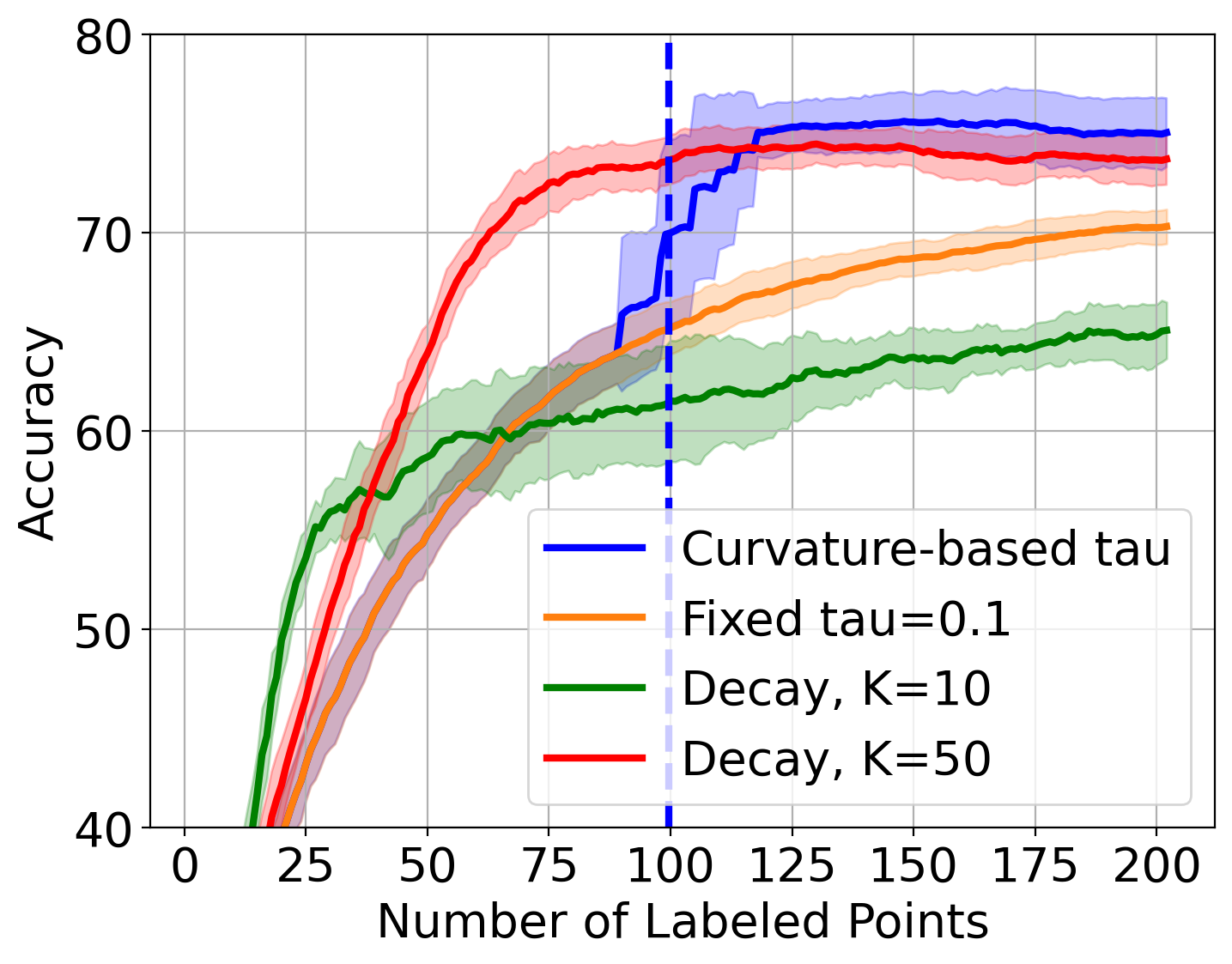}
    \caption{EMNIST}
    \end{subfigure}
    \hfill
    \begin{subfigure}[t]{0.45\linewidth}  
    \includegraphics[width=\linewidth]{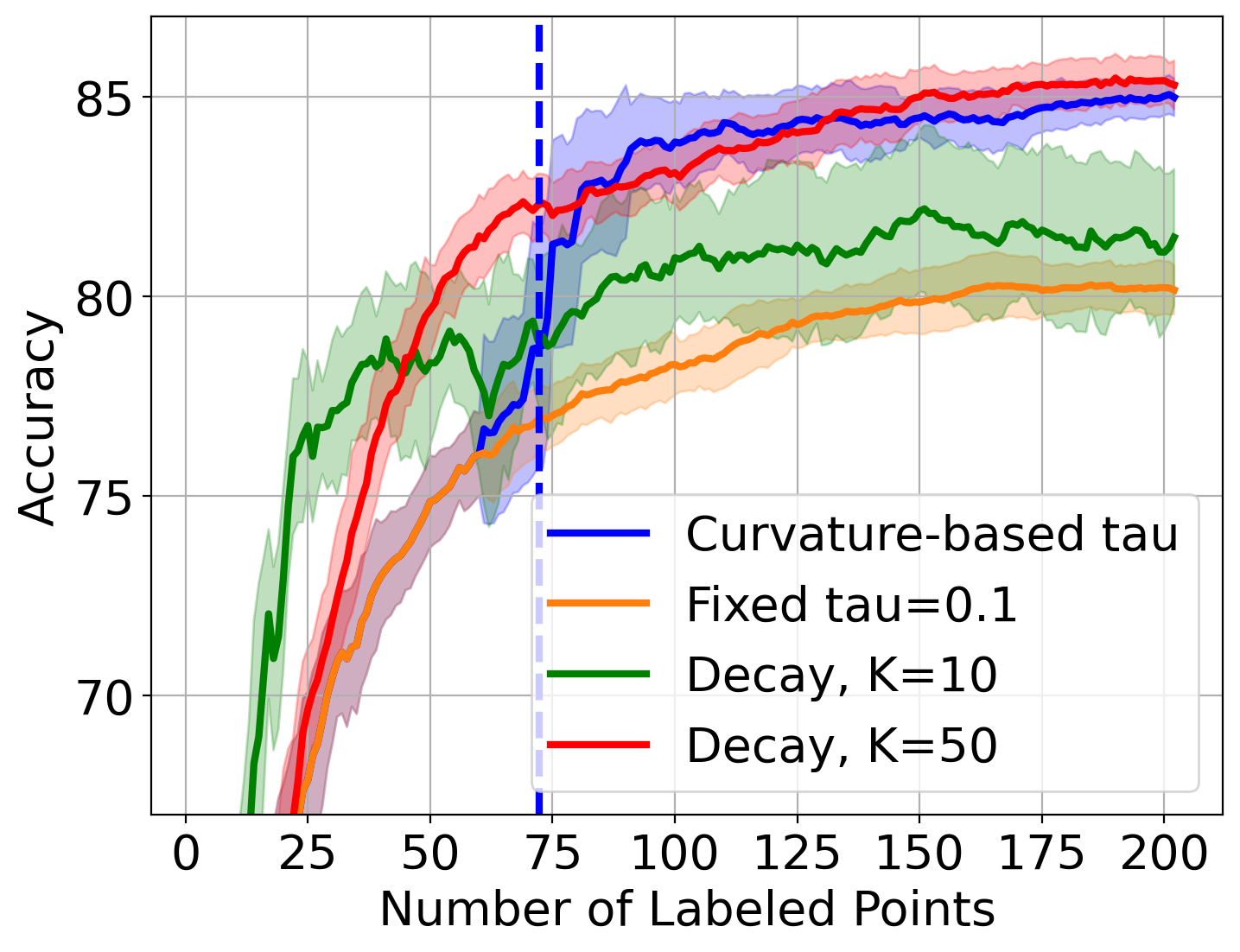}
    \caption{FashionMNIST}
    \end{subfigure}
    \hfill
    \begin{subfigure}[t]{0.45\linewidth}  
    \includegraphics[width=\linewidth]{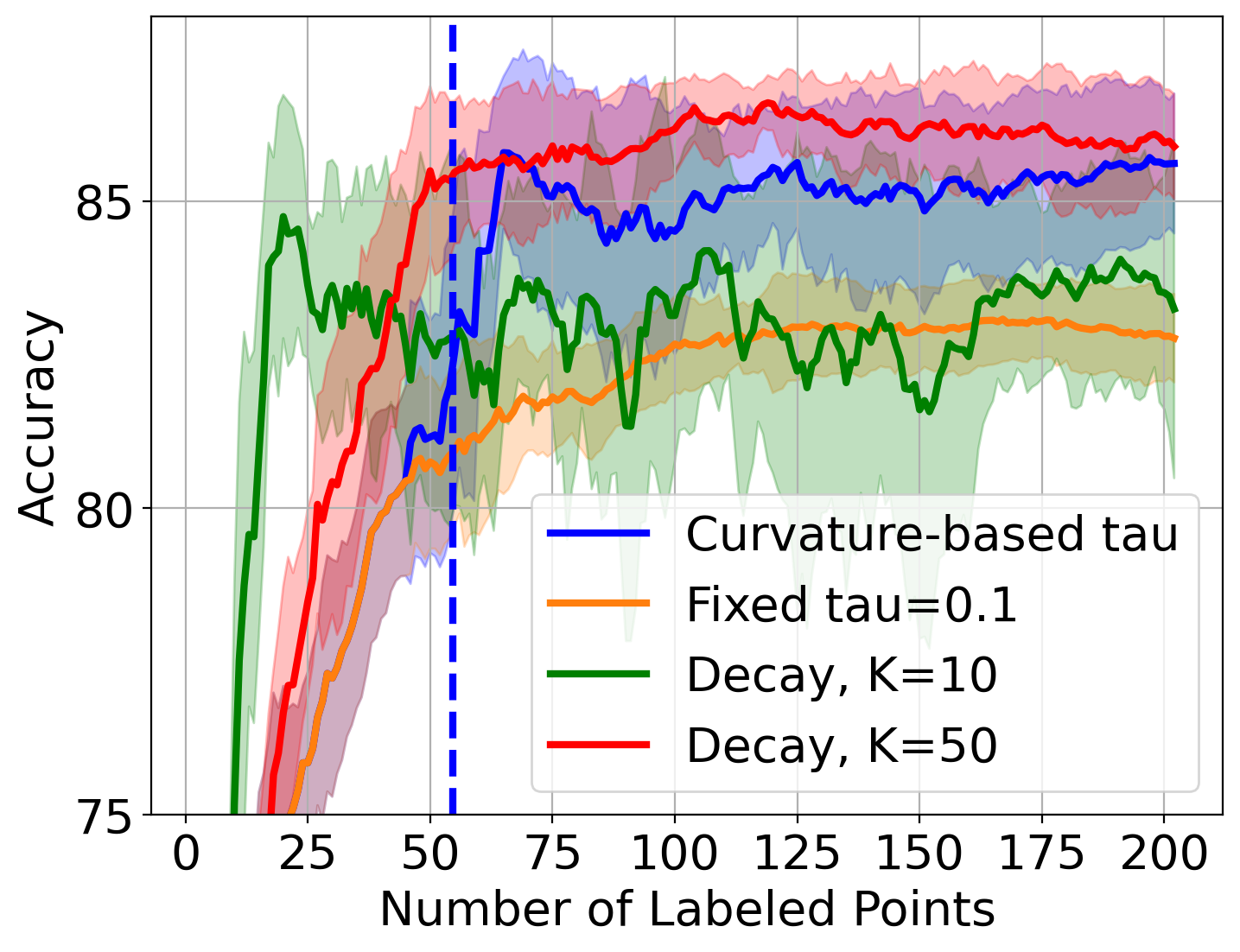}
    \caption{CIFAR-10}
    \end{subfigure}
    \hfill
    \begin{subfigure}[t]{0.45\linewidth}  
    \includegraphics[width=\linewidth]{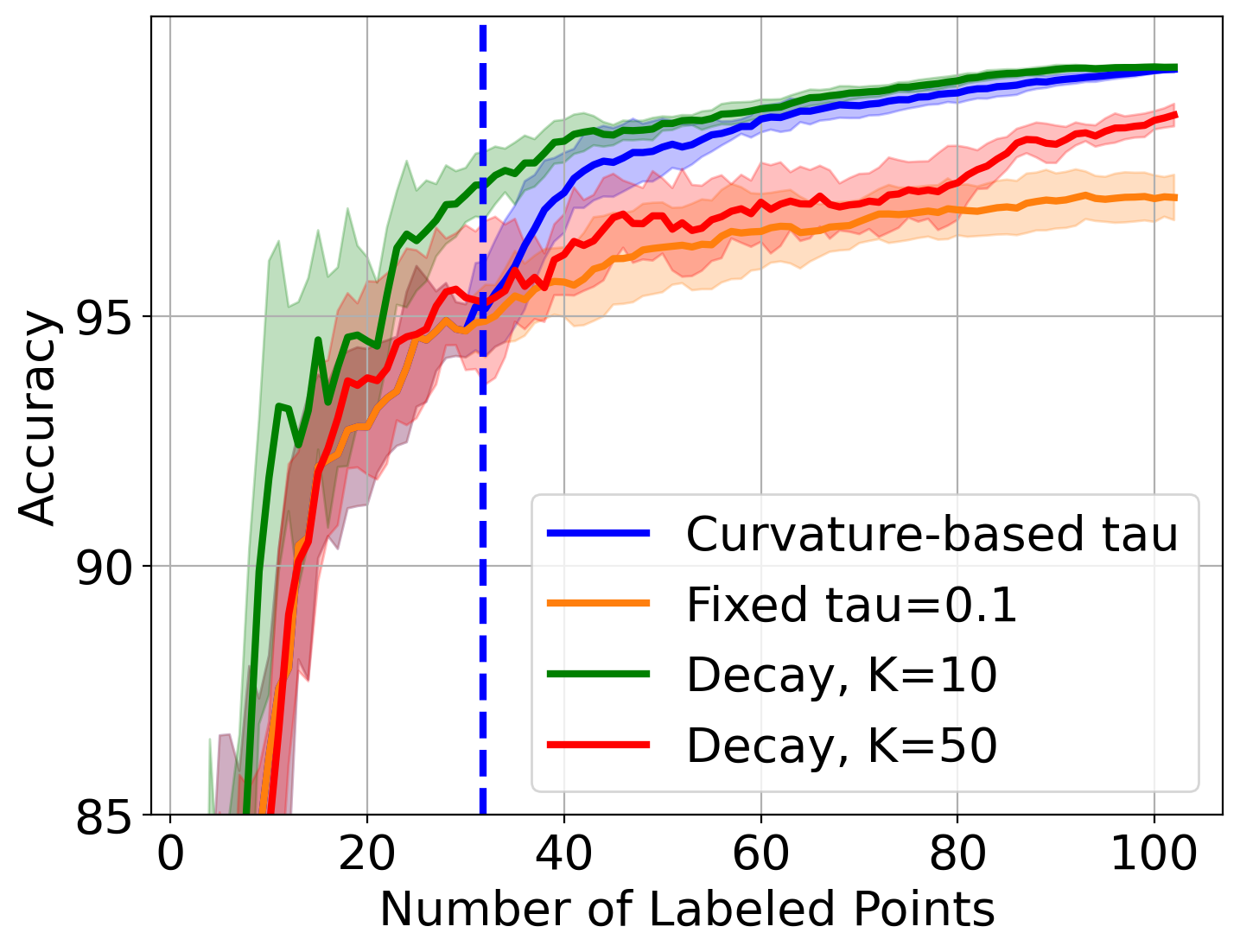}
    \caption{Box}
    \end{subfigure}
    \caption{Performance of PWLL-$\tau$ with the minimum norm uncertainty acquisition function under different decay schedules. 
    The dashed line indicates the point at which our curvature-based method sets $\tau = 0$.
    }
    \label{fig:pwll}
\end{figure*}

This section presents results on extending CC to provide a curvature-based signal for determining when PWLL-$\tau$ should transition from exploration to exploitation - that is, when to set $\tau = 0$. As discussed in Section \ref{sec:background:pwll}, given a user-defined parameter $K$, standard PWLL-$\tau$ decays $\tau$ to zero over 2K acquisition steps. After that, the algorithm becomes fully exploitative.

We report results on 4 datasets: EMNIST \cite{cohen2017emnist} (an extension of MNIST using letters, with 47 classes), FashionMNIST, CIFAR-10, and Box, a two-class toy dataset consisting of a $65 \times 65$ lattice of points on the unit square with a class boundary at $x=0.3$, also used in \cite{miller2023poisson}. 
We adopt the re-labeling scheme from \cite{miller2023poisson} for EMNIST, FashionMNIST, and CIFAR-10. Specifically, we map original labels to their values modulo 5 for EMNIST and modulo 3 for FashionMNIST and CIFAR-10, creating “modulo” classes (e.g., (0,3,6,9), (1,4,7), and (2,5,8) for FashionMNIST and CIFAR-10). Each experiment is initialized with one labeled example per “modulo” class (e.g., 3 total labels for FashionMNIST and CIFAR-10). 
This yields more complex class boundaries and ensures some original classes are initially unlabeled, \textit{requiring} the active learner to discover them through exploration. We construct the graphs with $k_1=25$, and run 10 trials.
In all experiments, we initialize with $\tau_0 = 0.1$ and compare: \begin{itemize}
    \item Fixed $\tau$ throughout,
    \item Decay schedule from \cite{miller2023poisson} for various $K$,
    \item Our curvature-based method for updating $\tau_0 = 0.1$ to $\tau = 0$.
\end{itemize}

Figure \ref{fig:pwll} shows that the optimal moment to switch from exploration to exploitation varies with the dataset and its classification structure, and that our data‑driven signal reliably identifies this point without requiring any preset decay schedule. This is clearly illustrated by comparing the Box and EMNIST results. 

On Box - a dataset with simple topology and classification structure - $K=10$ outperforms $K=50$ because the dataset requires less exploration before exploitation. Our method mirrors this behavior by transitioning at around 30 labeled points, matching $K=10$ in performance after both become fully exploitative. 

On the more complex EMNIST dataset, $K=50$ outperforms $K=10$ because more exploration is required. Our curvature-based method adapts to this complexity and ultimately outperforms both schedules after transitioning to exploitation. Interestingly, both our method and the $K=50$ decay schedule set $\tau = 0$ at \textit{the same point} - after 100 labels, on average.  Despite this, our method achieves better downstream exploitative AL results, illustrating that the quality of exploration - not just its duration - matters. This suggests that a sharp transition from pure exploration to pure exploitation can be more effective than gradually decaying, as reflected in the marked jump in accuracy - consistently observed across all datasets - once $\tau$ is set to zero.

These results underscore the effectiveness of our curvature-based $\tau$ scheduler, which provides a lightweight and principled alternative to the $K$-based decay schedules that may not generalize well. 
Our method only switches to exploitation once sufficient coverage has been achieved, requires no hyperparameters, and adapts automatically to the graph’s complexity, making it particularly suitable for real-world applications where class structure or label budgets are unknown in advance.

\subsection{Localized Graph Rewiring} \label{sec:results:rewiring}

\begin{figure*}[!ht]
\centering
    \begin{subfigure}[t]{.32\linewidth}
    \centering
    \includegraphics[width=\linewidth]{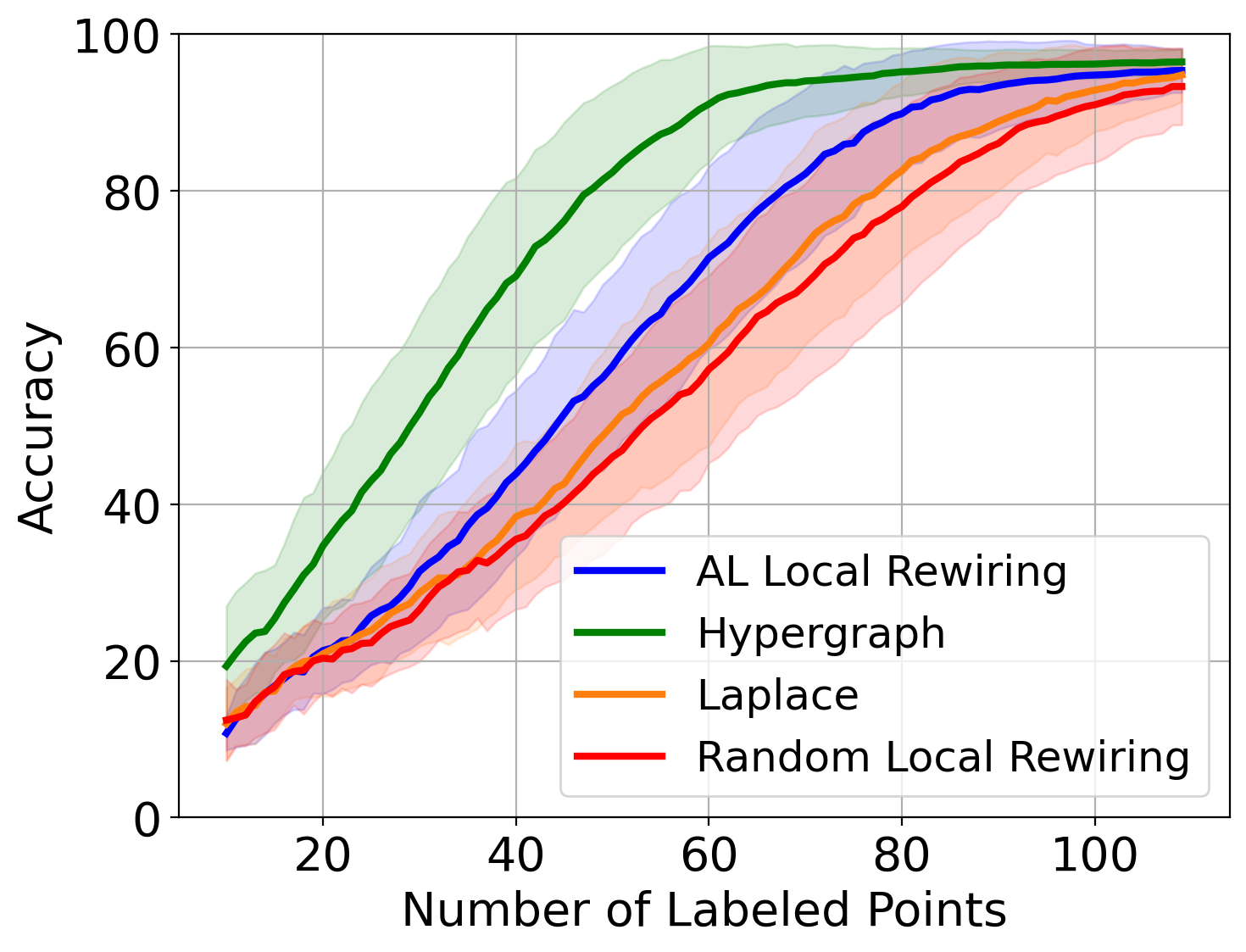}
    \caption{MNIST}
    \end{subfigure}
    \hfill
    \begin{subfigure}[t]{.32\linewidth}
    \centering
    \includegraphics[width=\linewidth]
    {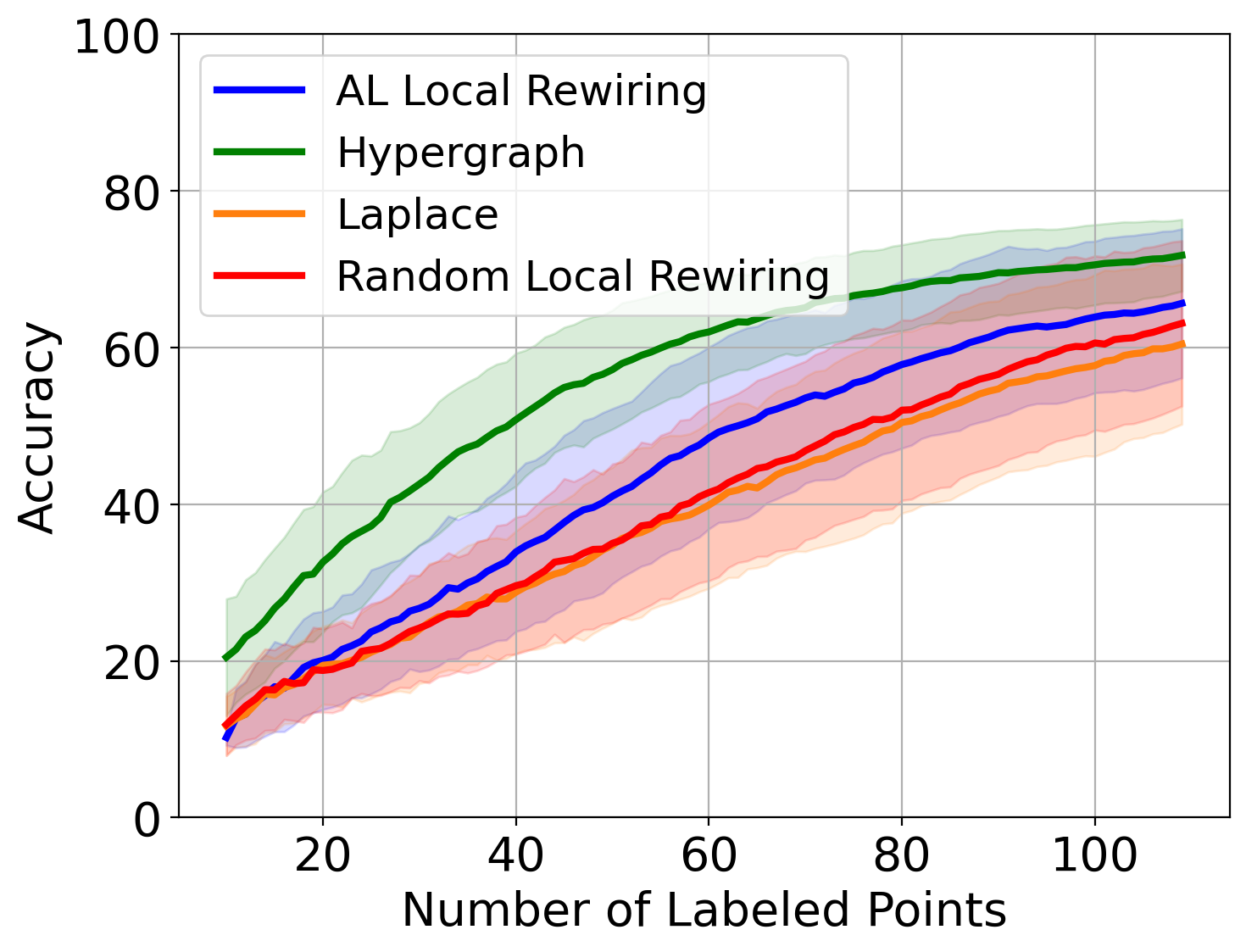}
    \caption{FashionMNIST}
    \end{subfigure}
    \hfill
    \begin{subfigure}[t]{.32\linewidth}
    \includegraphics[width=\linewidth]{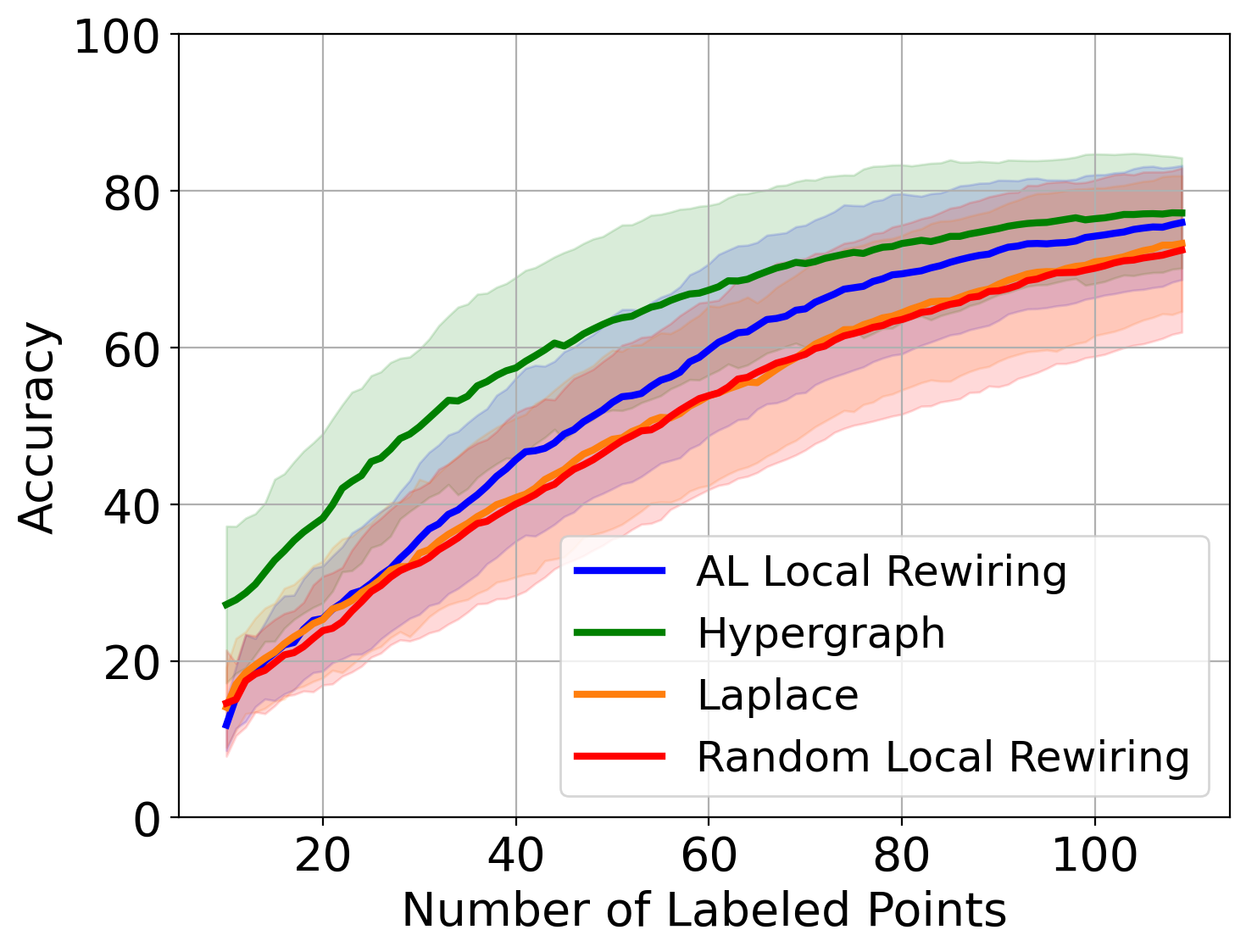}
    \caption{CIFAR-10}
    \end{subfigure}
    \hfill
    \caption{Comparison of sequential active learning performance with GBSSL variants on MNIST, FashionMNIST, and CIFAR-10. We start with 1 label per class and then run 100 AL iterations. We report mean (solid line) and standard deviation (shaded) over 100 trials. 
    }
    \label{fig:al_hypergraph}
\end{figure*}

\begin{table*}[t]
\centering
\footnotesize
\renewcommand{\arraystretch}{1.2}
\setlength{\tabcolsep}{6pt}
\begin{tabular}{|l|cc|cc|cc|}
\hline
\textbf{Method} &
\multicolumn{2}{c|}{\textbf{MNIST}} &
\multicolumn{2}{c|}{\textbf{FashionMNIST}} &
\multicolumn{2}{c|}{\textbf{CIFAR-10}} \\
\hline
 & Acc. (STD) & Time (s) & Acc. (STD) & Time (s) & Acc. (STD) & Time (s) \\
\hline
\textbf{AL-LR} & 69.84 (11.29) & 640 & 47.57 (11.75) & 731 & 58.76 (11.11) & 378\\
HG & 90.40 (7.70) & 9441 & 61.74 (6.40) & 10720 & 66.96 (11.09) & 4623\\
R-LR & 55.63 (12.69) &  627 & 40.96 (11.00) & 753 & 53.49 (12.21) & 416\\
Laplace & 59.38 (12.49) & 546 & 39.24 (10.40) & 656 & 53.14 (10.99) & 349\\
\hline
\end{tabular}
\caption{Accuracy and Time for 50 AL Iterations (60 total labeled points). We report means and standard deviations over 100 trials. The methods are AL Local Rewiring (AL-LR), Hypergraph (HG), Random Local Rewiring (R-LW), Laplace.}
\label{tab:al_hyper_50}
\end{table*}

\begin{table*}[t]
\centering
\footnotesize
\renewcommand{\arraystretch}{1.2}
\setlength{\tabcolsep}{6pt}
\begin{tabular}{|l|cc|cc|cc|}
\hline
\textbf{Method} &
\multicolumn{2}{c|}{\textbf{MNIST}} &
\multicolumn{2}{c|}{\textbf{FashionMNIST}} &
\multicolumn{2}{c|}{\textbf{CIFAR-10}} \\
\hline
 & Acc. (STD) & Time (s) & Acc. (STD) & Time (s) & Acc. (STD) & Time (s) \\
\hline
\textbf{AL-LR} & 95.37 (2.83) & 1273 & 65.65 (9.51) & 1439 & 75.96 (7.28) & 744 \\
HG & 96.47 (1.59) & 18376 & 71.77 (4.58) & 20340 & 77.18 (7.03) & 8908 \\
R-LR & 93.33 (4.86) & 1246 & 63.09 (10.55) & 1449 & 72.45 (10.48) & 821 \\
Laplace & 94.79 (3.34) & 1072 & 60.44 (10.23) & 1262 & 73.28 (8.65) & 683 \\
\hline
\end{tabular}
\caption{Accuracy and Time for 100 AL Iterations (100 trials). The methods are AL Local Rewiring (AL-LR), Hypergraph (HG), Random Local Rewiring (R-LW), Laplace.}
\label{tab:al_hyper}
\end{table*}

In this section, we evaluate the performance of our method outlined in Algorithm~\ref{alg:rewiring}.  
Our method, denoted AL Local Rewiring, is compared against three baselines: \begin{itemize}
    \item Hypergraph: the standard Hypergraph approach \eqref{eq:HOHL}, using full powers of the Laplacian on the entire graph,
    \item Laplace: using only the standard graph Laplacian $L_{k_1}$,
    \item Random Local Rewiring: a control strategy in which additional regularization is applied around randomly selected nodes, rather than the nodes acquired by the active learning policy.
\end{itemize}

We set $q = 2$, $k_1 = 50$, $k_2 = 30$, with powers $p_1 = 1$, $p_2 = 2$, and quadratic weights $\lambda_1 = 1$, $\lambda_2 = 4$, following the multiscale design of \cite{weihs2025Hypergraphs}.
Figure \ref{fig:al_hypergraph} visualizes the performance of each method throughout the AL process over 100 trials, and Tables \ref{tab:al_hyper_50} and \ref{tab:al_hyper} summarizes results and provides efficiency details.

The standard hypergraph approach, which applies full Laplacian powers over the entire graph, achieves the highest overall accuracy. This confirms the theoretical benefits of rich multiscale regularization when computational cost is not a limiting factor. Our proposed method, AL Local Rewiring, consistently ranks second and yields substantial improvements - typically 5–10\% over classical Laplace learning. These gains are consistent across all datasets and especially pronounced in the early rounds of active learning, where label scarcity hampers traditional methods. This demonstrates the effectiveness of localizing higher-order regularization around informative nodes. In contrast, the Random Local Rewiring control performs nearly identically to Laplace learning. This suggests that simply adding higher-order smoothing is not sufficient - targeted application around informative nodes is crucial.

In terms of efficiency, AL Local Rewiring matches the runtime of Laplace learning while avoiding the substantial overhead of the ``full'' hypergraph method, which is approximately ten times slower in our experiments. AL Local Rewiring therefore captures most of the performance benefits of full multiscale regularization at a fraction of the computational cost. Overall, our proposed strategy strikes the best balance of accuracy and efficiency by only computing and storing the higher order terms at active learning acquisitions. These results support the use of AL Local Rewiring as a scalable alternative to full multiscale regularization, particularly in real-world settings where efficiency is critical.

\section{Conclusion and Future Work} \label{sec:discussion}

This work introduces methodological and empirical advances that leverage graph-topological tools to improve both accuracy and efficiency in the low-label regime. First, we propose a novel coreset selection algorithm, Curvature Coreset (CC), which uses Balanced Forman Curvature to select representative labeled nodes that capture the graph’s underlying cluster structure. We further demonstrate that BFC serves as an effective signal for determining when exploration is complete, providing a principled stopping criterion and enhancing existing AL routines such as PWLL-$\tau$. Empirically, this leads to improved classifier initialization and consistently outperforms baseline strategies across multiple benchmark datasets in both sequential and batch AL settings. Together, these results establish curvature as a powerful mechanism for addressing the central \textit{exploration–exploitation trade-off} in graph-based AL.

Complementing this, a second focus of our work shows that \textit{modifying} the graph topology yields substantial performance gains. In particular, we introduce a localized multiscale regularization technique that selectively enhances structure around labeled nodes. This approach improves over standard Laplace learning in accuracy, while achieving speedups of over an order of magnitude compared to full hypergraph-based multiscale methods.

There are several promising directions for future work. First, a more principled understanding of how graph properties—such as homophily, degree distribution, or curvature—influence the exploration–exploitation trade-off could enable adaptive acquisition strategies tailored to specific graph regimes. For instance, we hypothesize that highly homophilous graphs may require fewer exploratory labels, while heterophilous graphs may benefit from broader initial coverage. “Heterophily-aware” models have already found success in graph attention networks \cite{wang2024heterophily}.
Second, curvature-based techniques open the door to new directions in large-scale and practical active learning settings. For batch active learning, curvature-driven alternatives to LocalMax \cite{chapman_novel_2023} - a method that extends the acquisition process to batches by relying solely on neighbor information - could better reflect the graph topology, enhancing data efficiency and ensuring balanced exploration across diverse regions of the graph. For graph reduction, aimed at enabling efficient learning on very large datasets, curvature-informed grouping criteria could yield geometrically consistent reductions while preserving key structural features.
Finally, our use of BFC highlights the untapped potential of integrating tools from the GNN literature into GBSSL. While GNNs excel in flexibility, scalability, and empirical performance on large datasets, GBSSL offers a mathematically tractable framework that can succeed with minimal labeled data. Techniques such as curvature-based metrics, rewiring strategies, and spectral objectives - typically used to improve message passing in GNNs - may have powerful analogues in GBSSL, offering new opportunities for both algorithm design and theoretical insight.
Exploring these connections could lead to unified frameworks that bridge geometric, spectral, and information-theoretic perspectives on label-efficient learning.

\section*{CRediT Authorship Contribution Statement}

\textbf{Harris Hardiman-Mostow:} Conceptualization, Methodology, Software, Validation, Formal analysis, Investigation, Writing - Original Draft, Writing - Review \& Editing, Visualization, Project administration, Funding acquisition. \textbf{Jack Mauro:} Conceptualization, Methodology, Software, Validation, Investigation, Writing - Original Draft, Writing - Review \& Editing, Visualization. \textbf{Adrien Weihs:} Conceptualization, Methodology, Formal analysis, Investigation, Writing - Original Draft, Writing - Review \& Editing, Supervision, Project administration. \textbf{Andrea L. Bertozzi:} Conceptualization, Resources, Writing - Review \& Editing, Supervision, Project administration, Funding acquisition.

\section*{Data Availability}

The datasets used in this work are either freely available online and cited in the paper manuscript, or available on our GitHub repository \url{https://github.com/hardiman-mostow/TopologyActiveLearning}.

\section*{Acknowledgments}

The authors thank James Chapman for helpful discussions about coresets. HHM was supported by the  National Science Foundation Graduate Research Fellowship Program under Grant No. DGE-2034835. ALB was supported by NSF grants DMS-2152717 and DMS-2318817 and the U.S. Department of Energy, Office of Science, Office of Advanced Scientific Computing Research, USA under award DE-SC0025589.  Any opinions, findings, and conclusions or recommendations expressed in this material are those of the authors and do not necessarily reflect the views of the National Science Foundation. This work used computational and storage services associated with the Hoffman2 Cluster which is operated by the UCLA Office of Advanced Research Computing’s Research Technology Group.

\section*{Declaration of Generative AI and AI-Assisted Technologies in the Manuscript Preparation Process}

During the preparation of this work the authors used ChatGPT and Claude for basic editing and proofreading. After using these tools, the authors reviewed and edited the content as needed and take full responsibility for the content of the published article.

\bibliography{references}

\begin{thebibliography}{10}
\expandafter\ifx\csname url\endcsname\relax
  \def\url#1{\texttt{#1}}\fi
\expandafter\ifx\csname urlprefix\endcsname\relax\def\urlprefix{URL }\fi
\expandafter\ifx\csname href\endcsname\relax
  \def\href#1#2{#2} \def\path#1{#1}\fi

\bibitem{iscen2019label}
A.~Iscen, G.~Tolias, Y.~Avrithis, O.~Chum, Label propagation for deep semi-supervised learning, in: Proceedings of the IEEE/CVF conference on computer vision and pattern recognition, 2019, pp. 5070--5079.

\bibitem{sellars2022laplacenet}
P.~Sellars, A.~I. Aviles-Rivero, C.-B. Sch{\"o}nlieb, Laplacenet: A hybrid graph-energy neural network for deep semisupervised classification, IEEE Transactions on Neural Networks and Learning Systems 35~(4) (2022) 5306--5318.

\bibitem{murphy2018unsupervised}
J.~M. Murphy, M.~Maggioni, Unsupervised clustering and active learning of hyperspectral images with nonlinear diffusion, IEEE Transactions on Geoscience and Remote Sensing 57~(3) (2018) 1829--1845.

\bibitem{brown2023utilizing}
J.~Brown, R.~O'Neill, J.~Calder, A.~L. Bertozzi, Utilizing contrastive learning for graph-based active learning of {SAR} data, in: Algorithms for Synthetic Aperture Radar Imagery XXX, Vol. 12520, SPIE, 2023, pp. 179--193.

\bibitem{enwright2023deep}
J.~Enwright, H.~Hardiman-Mostow, J.~Calder, A.~Bertozzi, Deep semi-supervised label propagation for {SAR} image classification, in: Algorithms for Synthetic Aperture Radar Imagery XXX, Vol. 12520, SPIE, 2023, pp. 158--170.

\bibitem{iyer2020graph}
G.~Iyer, J.~Chanussot, A.~L. Bertozzi, A graph-based approach for data fusion and segmentation of multimodal images, IEEE Transactions on Geoscience and Remote Sensing 59~(5) (2020) 4419--4429.

\bibitem{wang2021graph}
B.~Wang, S.~J. Osher, Graph interpolating activation improves both natural and robust accuracies in data-efficient deep learning, European Journal of Applied Mathematics 32~(3) (2021) 540--569.

\bibitem{liu2019learning}
Y.~Liu, J.~Lee, M.~Park, S.~Kim, E.~Yang, S.~J. Hwang, Y.~Yang, Learning to propagate labels: Transductive propagation network for few-shot learning, in: 7th International Conference on Learning Representations, ICLR 2019, International Conference on Learning Representations, ICLR, 2019.

\bibitem{huang2021ptn}
H.~Huang, J.~Zhang, J.~Zhang, Q.~Wu, C.~Xu, {PTN}: A {Poisson} transfer network for semi-supervised few-shot learning, in: Proceedings of the AAAI Conference on Artificial Intelligence, Vol.~35, 2021, pp. 1602--1609.

\bibitem{bertozziStuart}
A.~L. Bertozzi, X.~Luo, A.~M. Stuart, K.~C. Zygalakis, Uncertainty quantification in graph-based classification of high-dimensional data, SIAM/ASA Journal on Uncertainty Quantification 6~(2) (2018) 568--595.

\bibitem{weihs2025Hypergraphs}
A.~Weihs, A.~L. Bertozzi, M.~Thorpe, Analysis of semi-supervised learning on hypergraphs, arXiv preprint arXiv:2510.25354 (2025).

\bibitem{weihs2025HOHL}
A.~Weihs, A.~L. Bertozzi, M.~Thorpe, Higher-order regularization learning on hypergraphs, arXiv preprint arXiv:2510.26533 (2025).

\bibitem{chapman_novel_2023}
J.~Chapman, B.~Chen, Z.~Tan, J.~Calder, K.~Miller, A.~L. Bertozzi, Novel batch active learning approach and its application to synthetic aperture radar datasets, in: Algorithms for Synthetic Aperture Radar Imagery XXX, Vol. 12520, SPIE, 2023, pp. 95--110.

\bibitem{chen2024cgap}
B.~Chen, K.~Miller, A.~L. Bertozzi, J.~Schwenk, Cgap: A hybrid contrastive and graph-based active learning pipeline to detect water and sediment in multispectral images, IEEE Journal of Selected Topics in Applied Earth Observations and Remote Sensing (2024).

\bibitem{settles2009active}
B.~Settles, Active learning literature survey (2009).

\bibitem{miller2023poisson}
K.~Miller, J.~Calder, Poisson reweighted {Laplacian} uncertainty sampling for graph-based active learning, SIAM Journal on Mathematics of Data Science 5~(4) (2023) 1160--1190.

\bibitem{ricciBronstein}
J.~Topping, F.~D. Giovanni, B.~P. Chamberlain, X.~Dong, M.~M. Bronstein, Understanding over-squashing and bottlenecks on graphs via curvature, in: International Conference on Learning Representations, 2022.

\bibitem{Merkurjev}
E.~Merkurjev, D.~D. Nguyen, G.-W. Wei, Multiscale {Laplacian} learning, Applied Intelligence 53~(12) (2023) 15727--15746.

\bibitem{TutSpec}
U.~von Luxburg, \href{https://doi.org/10.1007/s11222-007-9033-z}{A tutorial on spectral clustering}, Statistics and Computing 17~(4) (2007) 395--416.
\newline\urlprefix\url{https://doi.org/10.1007/s11222-007-9033-z}

\bibitem{Slepcev}
D.~Slep{\v{c}}ev, M.~Thorpe, Analysis of $p$-{Laplacian} regularization in semisupervised learning, SIAM J. Math. Anal. 51~(3) (2019) 2085–2120.

\bibitem{calder_poisson_2020}
J.~Calder, B.~Cook, M.~Thorpe, D.~Slep{\v{c}}ev, Poisson learning: graph-based semi-supervised learning at very low label rates, in: Proceedings of the 37th {International} {Conference} on {Machine} {Learning}, PMLR, 2020, pp. 1306--1316, iSSN: 2640-3498.

\bibitem{LapRef}
X.~Zhu, Z.~Ghahramani, J.~Lafferty, Semi-supervised learning using gaussian fields and harmonic functions, in: Proceedings of the Twentieth International Conference on International Conference on Machine Learning, ICML'03, AAAI Press, 2003, p. 912–919.

\bibitem{NgSpectral}
A.~Ng, M.~Jordan, Y.~Weiss, On spectral clustering: Analysis and an algorithm, in: T.~Dietterich, S.~Becker, Z.~Ghahramani (Eds.), Advances in Neural Information Processing Systems, Vol.~14, MIT Press, 2001.

\bibitem{GARCIATRILLOS2018239}
N.~{Garcia Trillos}, D.~Slep{\v{c}}ev, A variational approach to the consistency of spectral clustering, Applied and Computational Harmonic Analysis 45~(2) (2018) 239--281.

\bibitem{hoffmann2020spectral}
F.~Hoffmann, B.~Hosseini, A.~A. Oberai, A.~M. Stuart, Spectral analysis of weighted {Laplacians} arising in data clustering, Applied and Computational Harmonic Analysis 56 (2022) 189--249.

\bibitem{rewiringMontufar}
K.~Karhadkar, P.~K. Banerjee, G.~Montufar, Fo{SR}: First-order spectral rewiring for addressing oversquashing in {GNN}s, in: The Eleventh International Conference on Learning Representations, 2023.

\bibitem{miller_model-change_2024}
K.~Miller, A.~L. Bertozzi, Model-change active learning in graph-based semi-supervised learning, Communications on Applied Mathematics and Computation 6~(2) (2024) 1270--1298.
\newblock \href {https://doi.org/10.1007/s42967-023-00328-z} {\path{doi:10.1007/s42967-023-00328-z}}.

\bibitem{ji12}
M.~Ji, J.~Han, A variance minimization criterion to active learning on graphs, in: N.~D. Lawrence, M.~Girolami (Eds.), Proceedings of the Fifteenth International Conference on Artificial Intelligence and Statistics, Vol.~22 of Proceedings of Machine Learning Research, PMLR, La Palma, Canary Islands, 2012, pp. 556--564.

\bibitem{ma2013sigma}
Y.~Ma, R.~Garnett, J.~Schneider, $\sigma$-optimality for active learning on gaussian random fields, Advances in Neural Information Processing Systems 26 (2013).

\bibitem{cloninger2021cautious}
A.~Cloninger, H.~N. Mhaskar, Cautious active clustering, Applied and Computational Harmonic Analysis 54 (2021) 44--74.

\bibitem{ollivier2010survey}
Y.~Ollivier, A survey of ricci curvature for metric spaces and markov chains, in: Probabilistic approach to geometry, Vol.~57, Mathematical Society of Japan, 2010, pp. 343--382.

\bibitem{feng_graph_nodate}
A.~Feng, M.~Weber, Graph pooling via ricci flow, Transactions on Machine Learning Research (2024).

\bibitem{tian2025curvature}
Y.~Tian, Z.~Lubberts, M.~Weber, Curvature-based clustering on graphs, Journal of Machine Learning Research 26~(52) (2025) 1--67.

\bibitem{zheng2023coveragecentric}
H.~Zheng, R.~Liu, F.~Lai, A.~Prakash, Coverage-centric coreset selection for high pruning rates, in: The Eleventh International Conference on Learning Representations, 2023.

\bibitem{ma2023homophilynecessitygraphneural}
Y.~Ma, X.~Liu, N.~Shah, J.~Tang, Is homophily a necessity for graph neural networks?, in: International Conference on Learning Representations, 2022.

\bibitem{mnist}
C.~J.~B. Yann~LeCun, Corinna~Cortes, {MNIST} handwritten digit database, \url{http://yann.lecun.com/exdb/mnist/} (2010).

\bibitem{xiao2017fashion}
H.~Xiao, K.~Rasul, R.~Vollgraf, Fashion-{MNIST}: a novel image dataset for benchmarking machine learning algorithms, arXiv preprint arXiv:1708.07747 (2017).

\bibitem{cifar}
A.~Krizhevsky, G.~E. Hinton, Learning multiple layers of features from tiny images, Tech. rep., University of Toronto, technical report (2009).

\bibitem{brown2024gll}
J.~Brown, B.~Chen, H.~Hardiman-Mostow, J.~Calder, A.~L. Bertozzi, {GLL}: A differentiable graph learning layer for neural networks, arXiv preprint arXiv:2412.08016 (2024).

\bibitem{miller_graph-based_2022}
K.~Miller, J.~Mauro, J.~Setiadi, X.~Baca, Z.~Shi, J.~Calder, A.~L. Bertozzi, Graph-based active learning for semi-supervised classification of {SAR} data, in: Algorithms for Synthetic Aperture Radar Imagery XXIX, Vol. 12095, SPIE, 2022, pp. 126--139.

\bibitem{vae}
D.~P. Kingma, M.~Welling, Auto-encoding variational bayes, in: 2nd International Conference on Learning Representations, {ICLR} 2014, Banff, AB, Canada, April 14-16, 2014, Conference Track Proceedings, 2014.

\bibitem{simclr}
T.~Chen, S.~Kornblith, M.~Norouzi, G.~Hinton, A simple framework for contrastive learning of visual representations, in: H.~D. III, A.~Singh (Eds.), Proceedings of the 37th International Conference on Machine Learning, Vol. 119 of Proceedings of Machine Learning Research, PMLR, 2020, pp. 1597--1607.

\bibitem{graphlearning}
J.~Calder, \href{https://doi.org/10.5281/zenodo.5850940}{Graphlearning python package} (Jan. 2022).
\newline\urlprefix\url{https://doi.org/10.5281/zenodo.5850940}

\bibitem{cohen2017emnist}
G.~Cohen, S.~Afshar, J.~Tapson, A.~Van~Schaik, {EMNIST}: Extending {MNIST} to handwritten letters, in: 2017 international joint conference on neural networks (IJCNN), IEEE, 2017, pp. 2921--2926.

\bibitem{wang2024heterophily}
J.~Wang, Y.~Guo, L.~Yang, Y.~Wang, Heterophily-aware graph attention network, Pattern Recognition 156 (2024) 110738.

\end{thebibliography}

\appendix
\section{Reduction Parameter}\label{sec:app:reduction}

\begin{figure}[!ht]
    \centering
    \includegraphics[width=.32\linewidth]{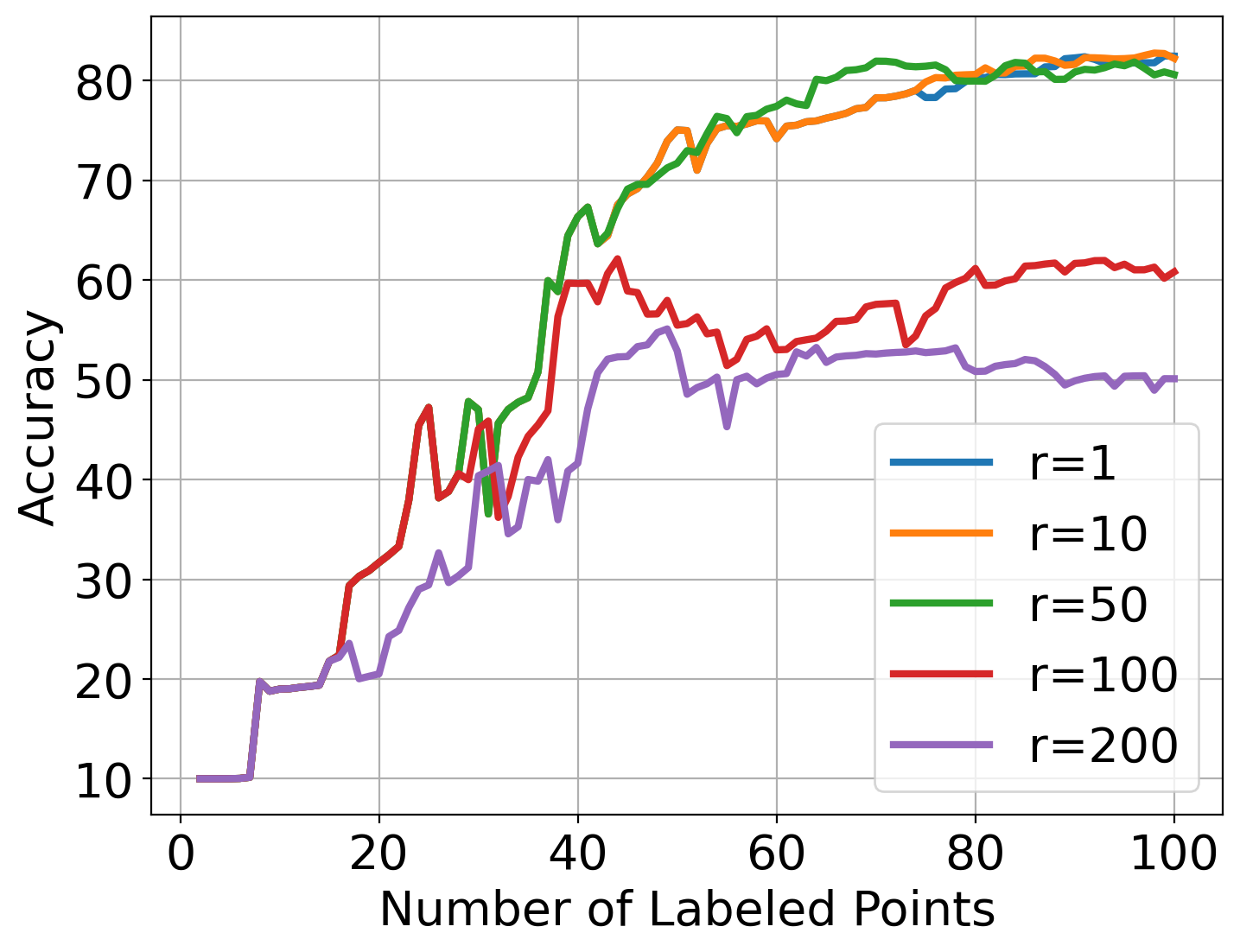}
    \caption{Comparison of reduction parameter $r$ values for CC on CIFAR-10. $r$ reduces the search space and helps speed up CC without sacrificing accuracy, up to a certain threshold where the search space becomes too small.}
    \label{fig:app:reduction_param}
\end{figure}

In order to speed up CC, we proposed a ``reduction parameter'' $r$, which reduces the search space by a factor of $r$ by only considering the top $1/r$ nodes by degree (see Remark \ref{rem:reduction}). This aligns with the definition of BFC (Definition \ref{def:bfc}) which is already biased toward high degree nodes. Figure \ref{fig:app:reduction_param} illustrates (on CIFAR-10) that this reduction parameter has no effect on the accuracy of CC, so long as $r$ is not too large. For example, $r=50$ provides almost identical accuracy, while being about $50$ times faster. Increasing $r$ beyond that appears to reduce the search space too much, causing worse performance. We also give timing details in Table \ref{tab:app:reduction_param_time}.

\begin{table}[!ht]
    \centering
    \begin{tabular}{||c|c|c|c|c|c||}
    \hline
    $r$ & 1 & 10 & 50 & 100 & 200 \\
    \hline
    Time (s) & 1908 & 195 & 44 & 25 & 15 \\
    \hline
    \end{tabular}
    \caption{The reduction parameter's effect on time to acquire 100 points on CIFAR-10.}
    \label{tab:app:reduction_param_time}
\end{table}

\section{DAC Coreset Sizes}\label{sec:app:dac_coreset_sizes}

One disadvantage of DAC is the high variability of coreset sizes. Due to the stochasticity of the algorithm, coreset sizes differ between trials even with a fixed radius $R$.
This means that DAC may produce drastically different results, even for the same data and the same radius. This also could pose problems in real-world settings where labeling budgets may be fixed. To illustrate this, and to provide transparency in our experiments, Table \ref{tab:dac_coreset_size} reports DAC coreset statistics on the image datasets. Even for a fixed radius, DAC produces drastically different coreset sizes. For example, on FashionMNIST, a radius of $0.4$ results in coreset sizes ranging from 89 to 159.

\begin{table*}[!ht]
\centering
\footnotesize
\begin{tabular}{lcccccc}
\hline
\textbf{Dataset} & \textbf{Target Coreset Size} & \textbf{Radius} & \textbf{Min} & \textbf{Mean} & \textbf{Max} & \textbf{Std}  \\
\hline
MNIST & 50  &  0.25  & 47 & 49.70 & 56 & 2.53 \\
MNIST & 100 & 0.125 & 96 & 104.80 & 140 & 12.13 \\
FashionMNIST & 50 & 0.4 & 47 & 53.60 & 76 & 7.88 \\
FashionMNIST & 100 & 0.2 & 89 & 102.80 & 159 & 21.26 \\
CIFAR-10 & 50 & 0.35 & 50 & 55.80 & 61 & 3.22 \\
CIFAR-10 & 100 & 0.175 & 92 & 100.50 & 104 & 3.44 \\
\hline
\end{tabular}
\caption{DAC Coreset Sizes Across 10 Trials}
\label{tab:dac_coreset_size}
\end{table*}

\end{document}